\documentclass[sn-mathphys-num]{sn-jnl}%

\usepackage{graphicx}%
\usepackage{multirow}%
\usepackage{amsmath,amssymb,amsfonts}%
\DeclareMathOperator*{\argmin}{arg\,min}
\usepackage{amsthm}%
\usepackage{mathtools}
\usepackage{mathrsfs}%
\usepackage[title]{appendix}%
\usepackage{xcolor}%
\usepackage{textcomp}%
\usepackage{manyfoot}%
\usepackage{booktabs}%
\usepackage{longtable,tabularx}
\usepackage{algorithm}%
\usepackage{algorithmicx}%
\usepackage{algpseudocode}%
\usepackage{listings}%
\usepackage{caption,subcaption}
\usepackage{xcolor} %
\usepackage[markup=underlined]{changes} %
\newcommand{\stkout}[1]{\ifmmode\text{\sout{\ensuremath{#1}}}\else\sout{#1}\fi}
\setdeletedmarkup{\stkout{#1}}

\setdeletedmarkup{\textcolor{red}{\stkout{#1}}}  %
\setaddedmarkup{{#1}}          %

\let\oldcite\cite
\renewcommand{\cite}[1]{\mbox{\oldcite{#1}}}
\let\old\deleted
\let\new\added

\renewcommand{\old}[1]{}     %
\renewcommand{\new}[1]{#1}   %

\newcommand{\Eqref}[1]{Eq.\nobreak\,\eqref{#1}}

\begin{document}

\title[Article Title]{Projection-based multifidelity linear regression for data-scarce
applications}

\author*[1]{\fnm{Vignesh} \sur{Sella}}\email{vignesh.sella@austin.utexas.edu}

\author[2]{\fnm{Julie} \sur{Pham}}\email{julie.pham@austin.utexas.edu}

\author[1,2,3]{\fnm{Karen} \sur{Willcox}}\email{kwillcox@oden.utexas.edu}

\author[1]{\fnm{Anirban} \sur{Chaudhuri}}\email{anirbanc@oden.utexas.edu}

\affil*[1]{\orgdiv{Oden Institute for Computational Engineering and Sciences}, \orgname{University of Texas at Austin}, \orgaddress{\postcode{78712}, \state{TX}, \country{USA}}}

\affil[2]{\orgdiv{Aerospace Engineering and Engineering Mechanics}, \orgname{University of Texas at Austin}, \orgaddress{\postcode{78712}, \state{TX}, \country{USA}}}

\affil[3]{\orgdiv{Santa Fe Institute}, \orgname{Santa Fe}, \orgaddress{\postcode{87501}, \state{NM}, \country{USA}}}

\abstract{Surrogate modeling for systems with high-dimensional quantities of interest remains challenging, particularly when training data are costly to acquire. This work develops multifidelity methods for multiple-input multiple-output linear regression targeting data-limited applications with high-dimensional outputs. Multifidelity methods integrate many inexpensive low-fidelity model evaluations with limited, costly high-fidelity evaluations. We introduce two projection-based multifidelity linear regression approaches \new{with linear and nonlinear features} that leverage principal component basis vectors for dimensionality reduction and combine multifidelity data through: (i) a direct data augmentation using low-fidelity data, and (ii) a data augmentation incorporating explicit linear corrections between low-fidelity and high-fidelity data. The data augmentation approaches combine high-fidelity and low-fidelity data into a unified training set and train the linear regression model through weighted least squares with fidelity-specific weights. \new{We introduce a proximity-based weighting scheme with automatic weight selection strategy through cross-validation.} The proposed multifidelity linear regression methods are demonstrated on approximating the surface pressure field of a hypersonic vehicle in flight \new{and the temperature field on an aircraft disc braking system}. In an ultra low-data regime of no more than \new{twelve} high-fidelity samples, multifidelity linear regression achieves approximately $\text{2\%}-\text{12\%}$ improvement in median accuracy \new{and a higher \textit{R}$^\text{2}$ score relative to single-fidelity methods at comparable computational cost.}}


\maketitle

\section{Introduction}\label{sec:intro}

An important challenge in scientific machine learning is to develop methods that can exploit and maximize the amount of learning possible from scarce data~\cite{baker2019basic,zhong2021machine,alzubaidi2023survey,carleo2019machine}. The need for such methods arises often in science and engineering, especially in the case of computational fluid dynamics (CFD), since expensive-to-evaluate high-fidelity (HF) models make many-query problems such as uncertainty quantification, risk analysis, optimization, and optimization under uncertainty computationally prohibitive~\cite{doi:10.1137/16M1082469}. Surrogate models that approximate the solutions to HF models can facilitate the design and analysis process; however, lack of sufficient HF data in tandem with high-dimensional quantities of interest adversely affect surrogate model accuracy. We propose multifidelity (MF) linear regression methods that leverage abundant low-cost, lower-fidelity (LF) data alongside severely limited HF data to construct linear regression models. These models operate within a reduced-dimensional subspace, obtained through the principal component analysis (PCA), to effectively handle both training data scarcity and the high dimensionality (on the order of tens of thousands of quantities of interest) inherent in our problem setting. 

Linear regression has been widely utilized as a surrogate modeling approach in aerospace applications due to its simplicity and interpretability. \new{We note that linear regression is a parametric surrogate modeling technique that encompasses a broad class of models that are \textit{linear in their parameters but can include features that are arbitrarily nonlinear functions of the input variables}~\cite{hastie2005elements}.} Traditionally, linear regression methods such as the response surface methodology (RSM) employ low-order polynomial approximations for problems characterized by a modest number of input variables (typically fewer than ten) and limited datasets ($\Theta(10^2)$ to $\Theta(10^3)$) due to computational costs~\cite{Madsen2000,Nakamura2022,Hosder2001,Mack2007}. In addition, many works have explored more data-intensive approaches, such as random forests or neural networks, that leverage significantly larger datasets, demonstrating strong predictive capabilities but requiring substantial computational resources~\cite{10.1145/2816795.2818129,4231016}. However, acquiring extensive HF training data often remains impractical for typical aerospace design applications without considerable computational investment. This motivates alternative approaches capable of working effectively in ultra low-data regime.

\new{MF linear regression techniques that leverage data of varying fidelity levels provide an effective means to reduce reliance on costly HF training data. Early MF approaches modeled the relationship between low- and high-fidelity outputs through additive or multiplicative correction factors applied to the LF model~\cite{balabanov1998multifidelity,doi:10.2514/3.10768,doi:10.2514/3.46462,doi:10.2514/2.2877}. These correction, or discrepancy, models were later extended using polynomial RSMs and other surrogate forms to capture more complex relationships~\cite{Madsen2001,vitali2002multi,eldred2004second,sun2011multi,zhou2015adaptive}. Broadly, such methods treat the HF response as a modified version of the LF prediction combined with an additional term that models the residual error~\cite{berci2014multidisciplinary,absi2020simulation,batra2019multifidelity}. Recent work has also explored linear (and polynomial) MF regression strategies that utilize ordinary least squares (OLS)-~\cite{zhang2018multifidelity} and moving least squares (MLS)-based methods~\cite{wang2021multi}. Both approaches incorporate scalar LF model outputs as either a basis function or a feature with additive or multiplicative correction terms. These correction-based methods are effective for scalar outputs, but their achievable model complexity is restricted by the number of available HF samples and dimensionality of the problem at hand. While in many engineering analyses only integrated or aggregate quantities (e.g., lift, drag, or maximum stress) are ultimately reported, access to the full-field response remains valuable. Full-field predictions provide flexibility for downstream tasks that can require different derived quantities or spatially localized information without the need to retrain or rerun simulations (or surrogates). They can enable identification of localized critical regions (e.g., areas of high pressure, stress, or temperature) that are important for tasks related to reliability, safety, and design optimization. In addition, full-field predictions can serve as inputs to coupled multi-physics analyses, where spatial fields of quantities such as pressure, velocity, or temperature are used as inputs.
Our work introduces a unified framework using weighted least squares (WLS) that combines LF and HF data in a single global regression with fidelity-informed weighting, enabling higher-order polynomial fits without the co-location or correction constraints of prior methods and can predict high-dimensional outputs of interest.}

\new{In the parametric multifidelity surrogate modeling context, recent studies on multifidelity neural networks have primarily focused on using larger quantities of data, often via discrepancy modeling, to improve estimation through a variety of machine learning techniques~\cite{MENG2020109020,ZHANG2021113485,sella2025improving,guo2022multi,conti2023multi}. These methods typically require datasets on the order of $\Theta(10^2)$–$\Theta(10^3)$ samples and are thus not applicable within the targeted sparse data regime. For the limited data setting, Bayesian nonparametric surrogate models, such as multifidelity Gaussian process regression or co-kriging, can be another surrogate modeling approach and have been investigated for low-dimensional output setting~\cite{forrester2007multi,le2014recursive}. However, given the cubic complexity with respect to the number of samples, Gaussian process regression can be computationally expensive as compared to linear regression when the training data contains numerous LF samples~\cite{williams1995gaussian}.}

In this work, we develop a MF linear regression method that can address regression problems involving high-dimensional outputs \new{with sparse training data.} The proposed method projects the outputs onto a lower-dimensional subspace obtained via PCA, facilitating effective modeling despite severely limited HF data. Note that the dimensionality of the input space in the applications considered in this work is significantly smaller than the output space. Our key methodological contribution is a data augmentation method that \new{integrates weighted least squares (WLS) with the dimensionality reduction and an automated weighting mechanism to explicitly incorporate LF data within a unified training process. We provide a closed-form expression for the projection-based MF linear regression coefficients following the derivation of the WLS solution.} Specifically, we introduce and analyze two methods for MF data augmentation: (i) direct data augmentation by combining fidelity sources, and (ii) data augmentation employing an explicit mapping between low- and HF outputs. \new{We present a proximity-based weighting scheme for WLS and an automatic weight selection strategy using cross-validation.} The WLS approach and adaptive data-driven weighting schemes enable appropriate utilization of LF data to improve predictive accuracy in high-dimensional, data scarce regimes. \new{We benchmark our proposed methods against state-of-the-art MF method with additive structure~\cite{zhang2018multifidelity,KOH_seminal}. We incorporate output dimensionality reduction into the additive MF method to provide a fair basis for comparison.} We demonstrate the effectiveness of the proposed MF methods through their application \new{on two engineering applications: predicting pressure load distributions on a hypersonic testbed vehicle and predicting the temperature field of aircraft disc brakes after a braking period.}

The remainder of this paper is structured as follows. Section \ref{sec:psetup} introduces the MF regression problem setup and the dimensionality reduction. Section \ref{sec:methods} details the proposed MF linear regression method using data augmentation \new{and an extension of the state-of-the-art additive MF regression strategy with projections is presented in Appendix~\ref{app:mf-add} for comparison. We present the numerical experiments on a hypersonic testbed vehicle application in Section \ref{sec:num_examples} and on an aircraft disc brake application in Section \ref{sec:num_example2}.} Finally, Section \ref{sec:conclusion} provides concluding remarks.

\section{Multifidelity regression: Background and problem formulation}
\label{sec:psetup}
We consider an MF regression setting involving training datasets obtained from models with different fidelity levels: an HF model that provides accurate predictions but is computationally expensive, and LF models that are computationally less costly but yield less accurate predictions.

\subsection{Background: Linear regression with dimensionality reduction}
Let the $d$-dimensional inputs to a system be denoted by $\boldsymbol{x} \in \mathcal{X} \subseteq \mathbb{R}^d$, where $\mathcal{X}$ is the input space, and the output quantity of interest be $\boldsymbol{y} \in \mathcal{Y} \subseteq \mathbb{R}^m$, defined on the output space $\mathcal{Y}$. In our target applications, $\boldsymbol{y}$ is a high-dimensional quantity, with $m$ typically on the order of tens of thousands. Due to the high-dimensionality of the outputs and limited HF data, we employ PCA to reduce the output dimension prior to regression. Similar projection-based approaches have been applied in the context of parametric reduced-order models~\cite{benner2015survey,SWISCHUK2019704,guo2019data}, as well as in neural network-based models~\cite{olearyroseberry2021derivativeinformed,sella2025improving}.

\textbf{Dimensionality reduction via PCA.}  For the training data matrix $\boldsymbol{Y} \in \mathbb{R}^{m\times N}$ with $N$ samples, the PCA basis vectors are obtained by standard PCA projection \cite{hastie2005elements}. We compute the principal components through the singular value decomposition (SVD). Given the row-wise sample mean of the training data $\boldsymbol{\bar{y}} \in \mathbb{R}^{m\times 1}$, we define $\overline{\boldsymbol{Y}} = \boldsymbol{\bar{y}}\boldsymbol{1}_N^\top$ where $\boldsymbol{1}_N \in \mathbb{R}^N$ is the length-$N$ column vector with all entries set to unity. For $N<m$, the thin SVD of the centered training data matrix is written as
\begin{equation}
  \boldsymbol{Y} - \overline{\boldsymbol{Y}} = \boldsymbol{U \Sigma V}^\top,
  \label{eq:svd}
\end{equation}
where $\boldsymbol{U}\in\mathbb{R}^{m\times N}$ and $\boldsymbol{V}\in\mathbb{R}^{N\times N}$ are orthogonal matrices, and $\boldsymbol{\Sigma}\in\mathbb{R}^{N\times N}$ is a diagonal matrix with non-decreasing entries of the singular values $\sigma_1 \geq \sigma_2 \geq \dots \geq \sigma_N \geq 0$. 
Given the left singular vectors $\boldsymbol{U}$, the reduced basis for projection to a lower-dimensional subspace of size $k\le N$ is the first $k$ columns $\boldsymbol{U}_k\in \mathbb{R}^{m \times k}$. The projection of the set of output samples $\boldsymbol{Y}$ on the low-dimensional subspace is given by the reduced states $\boldsymbol{C} \in \mathbb{R}^{k \times N}$, defined as
\begin{align}
    \label{eq:projection_down}
    \boldsymbol{C} = \boldsymbol{U}^\top_k(\boldsymbol{Y}-\overline{\boldsymbol{Y}}).
\end{align}
The dimension $k$ is chosen such that the cumulative variance captured by the first $k$ principal components is larger than a specified tolerance of $\epsilon$ as given by
\begin{align}
\label{eq:etol}
    \frac{\sum_{i=1}^{k} \sigma_i^2}{\sum_{i=1}^{N} \sigma_i^2} > \epsilon,
\end{align}
where $\sigma_i$ is the $i$-th singular value.

\textbf{Projection-based linear regression.} The regression problem considered in this work is a linear-regression-based surrogate model in the reduced-dimensional space $f: \mathbb{R}^d \to \mathbb{R}^k$, parameterized by regression coefficients $\boldsymbol\beta$. To address the high dimensionality of the output space, we perform regression in the reduced space defined by the first $k$ principal components obtained from PCA. Note that one could also apply dimensionality reduction to the inputs in addition to the outputs as shown by Sun~\cite{sun1996multivariate}. Alternate methods for dimensionality reduction in multivariate linear regression \cite{izenman1975reduced,li2003dimension} are also feasible and composable with the MF methods presented in the following section. \new{We employ PCA because it is interpretable, widely used, and has been shown to be effective across a broad range of applications~\cite{abdi2010principal}.}

We use training data projected to the reduced space using \Eqref{eq:projection_down} to obtain the projection-based linear regression model $f\left(\boldsymbol{x}; \boldsymbol\beta \right)$ for $k$-dimensional outputs as
\begin{align*}
    f\left(\boldsymbol{x}; \boldsymbol\beta \right) = \Phi(x)^{\top}\boldsymbol{\beta},
\end{align*}
where $\Phi(x)^{\top} \in \mathbb{R}^{p+1}$ is a $p+1$-dimensional feature vector that can include nonlinear transformations of the input (e.g., polynomial basis terms) and $\boldsymbol{\beta} \in \mathbb{R}^{p+1\times k}$ is the matrix of regression coefficients to be estimated. The surrogate model is therefore linear in the regression coefficients and can be trained using either ordinary or weighted least squares, depending on the MF regression methodology presented in the next section. We reconstruct the full-dimensional output from the regression predictions as
\begin{equation}
\label{eq:full_dim_output}
\widehat{\boldsymbol{y}}(\boldsymbol{x}^*) = \boldsymbol{U}_{k}f\left(\boldsymbol{x}^*; \boldsymbol\beta \right) + \overline{\boldsymbol{y}},    
\end{equation}
where $\widehat{\boldsymbol{y}}(\boldsymbol{x}^*)$ is the approximation of the true HF output for any new input $\boldsymbol{x}^* \in \mathcal{X}$.

\subsection{Multifidelity regression problem formulation} %
For ease of exposition, we consider a bifidelity setup, but the general idea can be extended to more than two fidelity levels. To distinguish between data originating from the HF and LF models, we define $\boldsymbol{X}_{\text{HF}} := \left[ \boldsymbol{x}^{(\text{HF})}_1, \dots, \boldsymbol{x}^{(\text{HF})}_{N_{\text{HF}}} \right] \in \mathbb{R}^{d \times N_{\text{HF}}}$ and $\boldsymbol{Y}_{\text{HF}} := \left[ \boldsymbol{y}^{(\text{HF})}_1, \dots, \boldsymbol{y}^{(\text{HF})}_{N_{\text{HF}}} \right] \in \mathbb{R}^{m \times N_{\text{HF}}}$ with analogous definitions for the LF data $\left(\boldsymbol{X}_{\text{LF}}, \boldsymbol{Y}_{\text{LF}}\right)$. In the applications of interest, we have $N_{\text{HF}} \ll N_{\text{LF}}$ and $N_{\text{HF}} \ll m$, reflecting the high computational cost of HF evaluations and the high-dimensionality of the output space.

The core supervised learning problem in this work is to construct a linear-regression-based surrogate model that accurately predicts the HF output $\boldsymbol{y}_{\text{HF}}(\boldsymbol{x}^*)$ for new inputs $\boldsymbol{x}^*$, by leveraging the bifidelity training dataset $(\boldsymbol{X}_{\text{HF}}, \boldsymbol{Y}_{\text{HF}})$ and $(\boldsymbol{X}_{\text{LF}}, \boldsymbol{Y}_{\text{LF}})$. The challenge of high output dimensionality ($m \gg N_{\text{HF}}$), and the limited number of HF samples makes direct regression in $\mathbb{R}^m$ ill-posed. Our approach mitigates these challenges by first projecting the HF and LF outputs to a lower-dimensional subspace using PCA, and then developing multifidelity regression methods in this reduced space as described in the following section.

\section{Projection-based multifidelity linear regression via data augmentation}
\label{sec:methods}

In this section, we develop a MF linear regression approach via data augmentation using the projected data. We first present two ways of synthetic data generation for data augmentation in Section~\ref{sec:DA} followed by the proximity-based weighting technique and the WLS approach developed for the MF linear regression in Section~\ref{sec:wls}. We then present an automated weight selection strategy through cross-validation in Section~\ref{sec:c-val}. 

\subsection{Synthetic data for data augmentation} \label{sec:DA}
The sparsity of HF data poses a significant challenge when attempting to fit more expressive surrogate models, such as polynomial regression with higher-order basis functions, because the HF dataset alone may not sufficiently constrain the model or allow for meaningful generalization. In contrast, training a surrogate model only on LF data is more feasible and less expensive, albeit at the cost of reduced accuracy. This work addresses these limitations by developing an MF linear regression method that utilizes an augmented training dataset, consisting of HF data and synthetic data derived from LF evaluations, to better constrain the regression model. The MF regression via data augmentation provides additional information about the underlying system response in regions of the input space insufficiently covered by HF samples. Furthermore, the MF approach facilitates the training of regression models with larger number of regression coefficients, for example, enabling the use of higher-degree polynomial bases beyond what the HF data alone would support.

Let $(\boldsymbol{X}_{\text{LF}}^\text{syn}, \boldsymbol{Y}_{\text{LF}}^\text{syn})$ denote the synthetic data used in data augmentation. We construct the synthetic data by one of two approaches:
\begin{enumerate}
    \item \textit{Direct augmentation}: The LF data are used directly as synthetic training data, i.e., $ (\boldsymbol{X}_{\text{LF}}^\text{syn}, \boldsymbol{Y}_{\text{LF}}^\text{syn}) = (\boldsymbol{X}_{\text{LF}}, \boldsymbol{Y}_{\text{LF}})$.
    \item \textit{Explicit mapping}: A learned linear correction map is applied to the LF data to approximate the HF behavior at $\boldsymbol{X}_{\text{LF}}$. This mapping is constructed by training a linear model $g$ between reduced-order representations of the LF and HF outputs in a shared low-dimensional space.
\end{enumerate}

The remainder of this section defines the explicit mapping approach, where we choose to model the relationship between LF and HF outputs in the subspace spanned by the reduced basis derived from the HF data as a linear transformation. We define a linear model $g: \mathbb{R}^k \to \mathbb{R}^k$ that maps reduced LF states to reduced HF states using the HF reduced basis $\boldsymbol{U}_k^\text{HF}$ to perform the dimensionality reduction. 

To train the model $g$, we need co-located HF and LF samples. When the LF samples are not co-located with the HF samples, we use a LF surrogate model to obtain the LF predictions at $\boldsymbol{X}_\text{HF}$. Let $f_\text{LF}: \mathbb{R}^d \to \mathbb{R}^k$ denote the linear regression surrogate model trained on the projected LF dataset $\left(\boldsymbol{X}_\text{LF}, \boldsymbol{C}_\text{LF}\right)$, where $\boldsymbol{C}_\text{LF}\in\mathbb{R}^{k\times N_\text{LF}}$ are the reduced LF states obtained via PCA as given by \Eqref{eq:projection_down}.
The predictions of reduced LF states at the HF input locations $\boldsymbol{X}_\text{HF}$ are obtained by $f_\text{LF}(\boldsymbol{X}_{\text{HF}})$ and reconstructed to the full-dimensional output space as $\boldsymbol{U}_k^\text{LF} f_\text{LF}(\boldsymbol{X}_{\text{HF}}) + \overline{\boldsymbol{Y}}_{\text{LF}}$ to obtain co-located LF output predictions.
Since the linear mapping $g$ operates within the subspace spanned by the HF reduced basis, we project the co-located LF predictions using the HF reduced basis to obtain the coordinate-transformed reduced LF states, $\widehat{\boldsymbol{C}}_\text{LF}$, as
\begin{equation} 
\label{eq:cob}    
\widehat{\boldsymbol{C}}_\text{LF}=\left(\boldsymbol{U}_k^\text{HF}\right)^\top \left( \left(\boldsymbol{U}_k^\text{LF} f_\text{LF}(\boldsymbol{X}_{\text{HF}}) + \overline{\boldsymbol{Y}}_{\text{LF}} \right) - \overline{\boldsymbol{Y}}_{\text{HF}}\right).
\end{equation}
The projection step in \Eqref{eq:cob} serves to express the LF predictions in the HF reduced basis.
Alternatively, one could explore methods such as manifold alignment to align the two subspaces and potentially provide better mappings between the two reduced states \cite{ham2005semisupervised,wang2009general}.
Simultaneously, we compute the HF reduced states as
\begin{align} 
\label{eq:reduced_HF_states}
\boldsymbol{C}_\text{HF}=\left(\boldsymbol{U}_k^\text{HF}\right)^\top \left(\boldsymbol{Y}_{\text{HF}} - \overline{\boldsymbol{Y}}_{\text{HF}}\right).
\end{align}
The linear mapping model $g$ is then trained via OLS on the co-located dataset $(\widehat{\boldsymbol{C}}_\text{LF},\boldsymbol{C}_\text{HF})$, where we are choosing to model this as a low-rank linear relationship between the LF and HF reduced outputs.
Note that if co-located data is already available, then one does not need to fit the LF surrogate model $f_\text{LF}$ and can directly obtain $\widehat{\boldsymbol{C}}_\text{LF}=\left(\boldsymbol{U}_k^\text{HF}\right)^\top \left( \boldsymbol{Y}_{\text{LF}}(\boldsymbol{X}_{\text{HF}}) - \overline{\boldsymbol{Y}}_{\text{HF}}\right)$ for training $g$.

Once trained, $g$ is used to generate synthetic data, $\boldsymbol{Y}_{\text{LF}}^\text{syn}$, at all the LF input locations $\boldsymbol{X}_{\text{LF}}$ by mapping the LF outputs as
\begin{align} \label{eq:synthetic_output_EM}
    \boldsymbol{Y}_{\text{LF}}^\text{syn} 
    = \boldsymbol{U}_k^\text{HF} g\left(\boldsymbol{C}_\text{LF}\right) + \overline{\boldsymbol{Y}}_{\text{HF}} 
    = \boldsymbol{U}_k^\text{HF} g\left(\left(\boldsymbol{U}_k^\text{LF}\right)^\top \left(\boldsymbol{Y}_{\text{LF}} - \overline{\boldsymbol{Y}}_{\text{LF}}\right) \right) + \overline{\boldsymbol{Y}}_{\text{HF}}.
\end{align}
This produces the synthetic dataset $(\boldsymbol{X}_{\text{LF}}^\text{syn}=\boldsymbol{X}_{\text{LF}}, \boldsymbol{Y}_{\text{LF}}^\text{syn})$ by explicit mapping, which is used for data augmentation in the MF regression method. We summarize this process in Alg.~\ref{alg:DA2}.
\begin{algorithm}[!htb]
\caption{Synthetic data generation via explicit linear mapping model} \label{alg:DA2}
\begin{algorithmic}[1]
\Statex \textbf{Input:} HF and LF training data $(\boldsymbol{X}_{\text{LF}}, \, \boldsymbol{Y}_{\text{LF}})$ and $(\boldsymbol{X}_{\text{HF}}, \, \boldsymbol{Y}_{\text{HF}})$
\Statex \textbf{Output:} Synthetic data $\boldsymbol{Y}_{\text{LF}}^\text{syn}$ at inputs $\boldsymbol{X}_{\text{LF}}$ from the LF to HF surrogate map
\State Project $\boldsymbol{Y}_{\text{LF}}$ to obtain the reduced states $\boldsymbol{C}_\text{LF}=\left(\boldsymbol{U}_k^\text{LF}\right)^\top \left(\boldsymbol{Y}_{\text{LF}} - \overline{\boldsymbol{Y}}_{\text{LF}}\right)$ \Comment{see \Eqref{eq:projection_down}}
\State Train the LF regression model $f_\text{LF}$ on $(\boldsymbol{X}_\text{LF},\boldsymbol{C}_\text{LF})$ using OLS
\State Generate co-located LF predictions as $\boldsymbol{U}_k^\text{LF} f_\text{LF}(\boldsymbol{X}_{\text{HF}}) + \overline{\boldsymbol{Y}}_{\text{LF}}$ at HF sample location $\boldsymbol{X}_{\text{HF}}$   
\State Project co-located LF predictions to obtain the coordinate-transformed LF reduced states $\widehat{\boldsymbol{C}}_\text{LF}$ via \Eqref{eq:cob}
\State Project $\boldsymbol{Y}_{\text{HF}}$ to obtain the reduced states $\boldsymbol{C}_\text{HF}$ using \Eqref{eq:reduced_HF_states}
\State Train $\text{LF}\mapsto \text{HF}$ linear regression model $g$ on $(\widehat{\boldsymbol{C}}_\text{LF},\boldsymbol{C}_\text{HF})$ using OLS
\State Generate synthetic data $\boldsymbol{Y}_{\text{LF}}^\text{syn}$ at $\boldsymbol{X}_{\text{LF}}$ locations using \Eqref{eq:synthetic_output_EM}
\end{algorithmic}
\end{algorithm}

\subsection{Weighted least squares using proximity-based weights} \label{sec:wls}
Given an augmented training dataset incorporating synthetic LF-derived samples, we train the MF surrogate model using weighted least squares regression to account for fidelity-dependent variance. Ordinary least squares (OLS) assumes homoscedasticity, or constant variance in the residuals, which does not hold in this setting, as synthetic samples derived from LF data are known \textit{a priori} to be a less accurate approximation. To account for this expected heteroscedasticity, we instead apply WLS~\cite{montgomery2021introduction} with distinct weights assigned to HF and synthetic training samples. Specifically, we define a diagonal weight matrix weight matrix $\boldsymbol{W}= \operatorname{diag}(w_1, \hdots, w_{N_{\text{HF}}+N_{\text{LF}}})$, where weights are assigned as
\begin{equation}\label{eq:weights}
w_i =
\begin{cases}
1, & i = 1, \dots, N_{\text{HF}} \\ 
h(w_{\text{syn}}) < 1 , & i = N_{\text{HF}} + 1, \dots, N_{\text{HF}} + N_{\text{LF}},
\end{cases}
\end{equation}
where $h(w_{\text{syn}})$ is a weighting function for LF training samples defined using the hyperparameters $w_{\text{syn}}$.

\new{In the context of MF surrogate modeling, the HF model is regarded as the best available source of ground truth for a given quantity of interest. Thus,} when applying least-squares linear regression, LF samples located near HF samples in the input space can be considered redundant or uninformative\new{,} since continuity ensures that proximity in the input space yields proximity of the outputs. \new{Consequently, the} LF data may introduce noise rather than useful information due to their inherent lower fidelity. This issue is particularly relevant when LF and HF datasets are fixed, which is the setting considered in this paper. In this context, the LF data introduces position-dependent variance, an instance of heteroscedasticity\new{, which is precisely what WLS was designed to handle~\cite{montgomery2021introduction}.} To mitigate this effect, we introduce a \textit{proximity-based weighting scheme} that down-weights LF samples located near HF samples. The sample weight assigned to a given LF point depends on (1) its distance to the nearest HF point and (2) whether it originates from the LF or HF source. This approach allows the model to emphasize LF samples that fill gaps (alleviate epistemic uncertainty) in the HF dataset while discounting those that are likely redundant. We compare the proximity-based weighting scheme with a fixed weighting scheme. The weighting function is then defined as
\begin{equation}\label{eq:weighting_function}
h(w_{\text{syn}})=
\begin{cases}
    w_{\text{syn}} & \text{fixed weights}\\
    \sigma \left (\rho(\boldsymbol{x}^{\text{LF}}, \boldsymbol{x}^{\text{HF}}); w_{\text{syn}} \right) & \text{proximity-based weights},
\end{cases}
\end{equation}
where $\rho : \mathbb{R}^d \times \mathbb{R}^d \to \mathbb{R}$ is a distance function (e.g., Euclidean distance) and $\sigma : \mathbb{R} \to [0, 1]$ is a monotonic transformation that maps distances to normalized weights. Suitable choices include Heaviside step functions, sigmoids, or any other similar function. In this work, we use a Heaviside step function to define $\sigma \left (\rho (\boldsymbol{x}^{\text{LF}}, \boldsymbol{x}^{\text{HF}}); w_{\text{syn}} \right)= w_{\text{syn}}\mathbf{1}_{\rho (\boldsymbol{x}^{\text{LF}}, \boldsymbol{x}^{\text{HF}})\ge \tau}$, where $\mathbf{1}$ is an indicator function that sets the maximum value to $w_\text{syn}$ and the minimum value to 0 depending on whether the distance from HF samples exceeds a threshold value of $\tau$. We use Euclidean distance as the distance function $\rho(\cdot)$. The value of $w_{\text{syn}}$ significantly impacts model performance and is selected via cross-validation, as described in Section~\ref{sec:c-val}.

The surrogate is trained in the projected output space defined by the HF reduced basis (see Section~\ref{sec:psetup}). The MF linear regression model denoted by $f_\text{MF}: \mathbb{R}^d \to \mathbb{R}^k$ is given by 
\begin{equation}
    f_\text{MF}(\boldsymbol{x}; w_\text{syn}) = \Phi(\boldsymbol{x})^\top \boldsymbol{\hat{\beta}}_\text{MF}(w_\text{syn}),
\end{equation}
where $\boldsymbol{\hat{\beta}}_\text{MF}(w_\text{syn})$ are the regression coefficients and has the explicit dependence on the synthetic sample weight since they are estimated using WLS.
The regression model $f_\text{MF}$ is trained on the augmented training dataset containing $N_\text{HF} + N_\text{LF}$ samples given by $([\boldsymbol{X}_{\text{HF}},\boldsymbol{X}_{\text{LF}}^\text{syn}], [\boldsymbol{Y}_{\text{HF}},\boldsymbol{Y}_{\text{LF}}^\text{syn}])$ as defined in Section~\ref{sec:DA}. For brevity, when utilizing the data augmentation method, we define $\boldsymbol{X}_\text{MF} := [\boldsymbol{X}_\text{HF}, \boldsymbol{X}_\text{LF}^{\text{syn}}]$ as the independent variables. 
Similarly, we define $\boldsymbol{Y}_{\text{MF}} := [\boldsymbol{Y}_{\text{HF}},\boldsymbol{Y}_{\text{LF}}^\text{syn}]$. Projecting these outputs yields reduced states,
\begin{align}\label{eq:reduced_MF_states}    \boldsymbol{C}_\text{MF}=\left(\boldsymbol{U}_k^\text{HF}\right)^\top \left(
\boldsymbol{Y}_{\text{MF}}
- \overline{\boldsymbol{y}}_{\text{HF}}\boldsymbol{1}_{N_\text{HF} + N_\text{LF}}^\top\right).
\end{align}

The optimal regression coefficients when the MF linear regression model is trained on $(\boldsymbol{X}_\text{MF},\boldsymbol{C}_\text{MF})$ using WLS with weights $\boldsymbol{W}$ can be obtained in closed-form as 
\begin{align}
\label{eq:WLS_beta}
    \boldsymbol{\hat{\beta}}^*_\text{MF}(w_\text{syn}) &=\left(\Phi\left(\boldsymbol{X_\text{MF}}\right)^\top\boldsymbol{W}\Phi\left(\boldsymbol{X_\text{MF}}\right)\right)^{-1}\Phi\left(\boldsymbol{X_\text{MF}}\right)^\top\boldsymbol{W}\boldsymbol{C}_\text{MF}  \\
    &= \left(\Phi\left(\boldsymbol{X_\text{MF}}\right)^\top\boldsymbol{W}\Phi\left(\boldsymbol{X_\text{MF}}\right)\right)^{-1}\Phi\left(\boldsymbol{X_\text{MF}}\right)^\top\boldsymbol{W}\left(\boldsymbol{U}_k^\text{HF}\right)^\top\left(\boldsymbol{Y}_{\text{MF}}- \overline{\boldsymbol{y}}_{\text{HF}}\boldsymbol{1}_{N_\text{HF} + N_\text{LF}}^\top\right) \label{eq:app_beta},
\end{align}
where the derivation for the closed-form expression in \Eqref{eq:WLS_beta} follows from the known WLS solution~\cite{hastie2005elements} and \Eqref{eq:app_beta} substitutes the reduced states.
The prediction at any new input location $\boldsymbol{x}^*$ is made in the reduced space and then lifted to the full-dimensional output space as
\begin{align}\label{eq:data_aug_output}
\begin{split}   
    \widehat{\boldsymbol{y}}_{\text{MF}}(\boldsymbol{x}^*; w_\text{syn}) &= \boldsymbol{U}_k^\text{HF}f_{\text{MF}}(\boldsymbol{x}^*; w_\text{syn}) + \overline{\boldsymbol{Y}}_{\text{HF}} \\
    &= \boldsymbol{U}_k^\text{HF}\Phi(\boldsymbol{x}^*)^\top \boldsymbol{\hat{\beta}}^*_\text{MF}(w_\text{syn}) + \overline{\boldsymbol{y}}_{\text{HF}}\boldsymbol{1}_{N_\text{HF} + N_\text{LF}}^\top 
    \end{split}\\
    &= \boldsymbol{U}_k^\text{HF}\Phi(\boldsymbol{x}^*)^\top \underbrace{\left(\Phi(\boldsymbol{X}_{\text{MF}})^\top\boldsymbol{W}\Phi(\boldsymbol{X}_{\text{MF}})\right)^{-1}\Phi(\boldsymbol{X}_{\text{MF}})^\top\boldsymbol{W}\boldsymbol{C}_\text{MF}}_{\boldsymbol{\hat{\beta}}^*_\text{MF}(w_\text{syn})} + \overline{\boldsymbol{y}}_{\text{HF}}\boldsymbol{1}_{N_\text{HF} + N_\text{LF}}^\top. \label{eq:wls_closed}
\end{align}
We summarize the data augmentation method for MF linear regression in Alg. \ref{alg:DA}. 
\begin{algorithm}[!htb]
\caption{Multifidelity linear regression via data augmentation} \label{alg:DA}
\begin{algorithmic}[1]
\Statex \textbf{Input:} HF and LF training data $(\boldsymbol{X}_{\text{LF}}, \, \boldsymbol{Y}_{\text{LF}})$ and $(\boldsymbol{X}_{\text{HF}}, \, \boldsymbol{Y}_{\text{HF}})$, synthetic sample weighting parameter $w_\text{syn}$, new input location for prediction $\boldsymbol{x}^*$
\Statex \textbf{Output:} Output predictions $\widehat{\boldsymbol{y}}_{\text{MF}}$ at inputs $\boldsymbol{x}^*$ from MF surrogate
\State Generate synthetic data by transforming the LF data: $(\boldsymbol{X}_{\text{LF}}, \, \boldsymbol{Y}_{\text{LF}}) \mapsto (\boldsymbol{X}_{\text{LF}}^\text{syn}, \boldsymbol{Y}_{\text{LF}}^\text{syn})$ \Comment{use Alg.~\ref{alg:DA2} for the explicit mapping method}
\State Augment the training dataset to contain $N_\text{HF}+N_\text{LF}$ samples: $([\boldsymbol{X}_{\text{HF}},\boldsymbol{X}_{\text{LF}}^\text{syn}], [\boldsymbol{Y}_{\text{HF}},\boldsymbol{Y}_{\text{LF}}^\text{syn}])$
\State Project $[\boldsymbol{Y}_{\text{HF}},\boldsymbol{Y}_{\text{LF}}^\text{syn}]$ to obtain the reduced states of MF training data outputs $\boldsymbol{C}_\text{MF}$ using \Eqref{eq:reduced_MF_states}
\State Set up sample weight matrix $\boldsymbol{W}$ based on choice of sample weighting scheme using Eqs.~\eqref{eq:weights} and \eqref{eq:weighting_function} 
\State Train MF linear regression surrogate model $f_{\text{MF}}$ on $([\boldsymbol{X}_{\text{HF}},\boldsymbol{X}_{\text{LF}}^\text{syn}],\boldsymbol{C}_\text{MF})$ with weights $\boldsymbol{W}$ using WLS \Comment{closed-form expression in \Eqref{eq:app_beta}}
\State Predict $\widehat{\boldsymbol{y}}_{\text{MF}}(\boldsymbol{x}^*)$ by reconstructing the output of $f_{\text{MF}}(\boldsymbol{x}^*;w_\text{syn})$ in the full-dimensional space defined in \Eqref{eq:data_aug_output}
\end{algorithmic}
\end{algorithm}

\subsection{Cross-validation for optimal sample weight selection}
\label{sec:c-val}
The synthetic sample weighting function in \Eqref{eq:weighting_function} has a tunable hyperparameter $w_\text{syn} \in (0,1)$. \new{Rather than arbitrarily selecting $w_\text{syn}$ or utilizing a heuristic-based approach,} we select $w_\text{syn}$ for the proximity-based weighting scheme by minimizing the prediction error using leave-one-out cross-validation (LOOCV). \new{LOOCV was chosen rather than general $k$-fold cross-validation since we are working with a limited number of HF samples.} For each HF training sample $i \in \{1,\dots,N_{\text{HF}}\}$, a model $f_{\text{MF}}(.; w_\text{syn})$ is trained on the remaining data and the validation error for the held-out sample is defined as
\begin{equation}
    \epsilon_{_\text{LOOCV}}\left(\boldsymbol{y}^\text{HF}_i; w_\text{syn}\right) := \frac{ \left\lVert \boldsymbol{y}^\text{HF}_i - \hat{\boldsymbol{y}}^\text{MF}_i(w_\text{syn}) \right\rVert_{2} }{ \left\lVert \boldsymbol{y}^\text{HF}_i \right\rVert_{2}},
    \label{eq:loocv_error}
\end{equation}
where $\hat{\boldsymbol{y}}^\text{MF}_i(w_\text{syn})$ denotes the prediction at $\boldsymbol{x}^\text{HF}_i$ made by the model trained without sample $i$.
The optimal weight hyperparameter $w^*_\text{syn}$ minimizes the mean LOOCV error over the HF training set as given by
\begin{equation}\label{eq:w_syn_loocv}    
w^*_\text{syn} = \argmin_{w_\text{syn}\in (0,1)} \frac{1}{N_\text{HF}} \sum_{i=1}^{N_\text{HF}} \epsilon_{_\text{LOOCV}}\left(\boldsymbol{y}_{i}^{\text{HF}};w_\text{syn}\right),
\end{equation} 
where $\epsilon_{_\text{LOOCV}}(\cdot;w_\text{syn})$ is the error function defined in \eqref{eq:loocv_error}. The optimization in \Eqref{eq:w_syn_loocv} is performed using the BFGS algorithm~\cite{doi:10.1137/0916069}. As shown in Section~\ref{sec:re_mf}, this procedure is critical for the robust performance of the data augmentation methods with proximity-based weighting, which are sensitive to the choice of $w_\text{syn}$.

\section{Numerical demonstration: hypersonic aerodynamics application}
\label{sec:num_examples}
In this section, we present the results for a hypersonic testbed vehicle problem in the CFD domain described in Section~\ref{sec:hype}. The HF and the LF models used for the MF linear regression are described in Section~\ref{sec:fidelity}. Then, we present results for the projection-based MF linear regression methods proposed in this work in Section~\ref{sec:re_mf}. \new{We compare against the state-of-the-art MF regression method using the additive approach presented} in~\cite{zhang2018multifidelity, KOH_seminal}\new{. We provide a fair comparison by combining the additive MF method} with dimensionality reduction of the outputs (see Appendix~\ref{app:mf-add}), and compare it against the MF methods presented in Section~\ref{sec:methods}.

\subsection{IC3X hypersonic vehicle problem description}\label{sec:hype}
In order to gain design insights for performance, stability, and reliability of a hypersonic vehicle, CFD simulations are required over a range of flight conditions. For example, stability analyses for a hypersonic vehicle require an understanding of the surface pressure field as a function of the operating flight conditions, namely, the Mach number, angle of attack, and sideslip angle of the vehicle. However, HF CFD solutions are computationally intensive due to the fine mesh size required to adequately capture the physics of hypersonic flight. In this work, we address the prohibitive computational cost through constructing cheaper approximations using MF linear regression techniques that reduce the number of HF model evaluations required to make accurate predictions of the pressure fields over a range of operating conditions by introducing data from cheaper LF models.

\begin{figure}[!htb]
	\centering
	\includegraphics[width = 5in, trim = 0cm 5cm 5cm 5cm, clip]{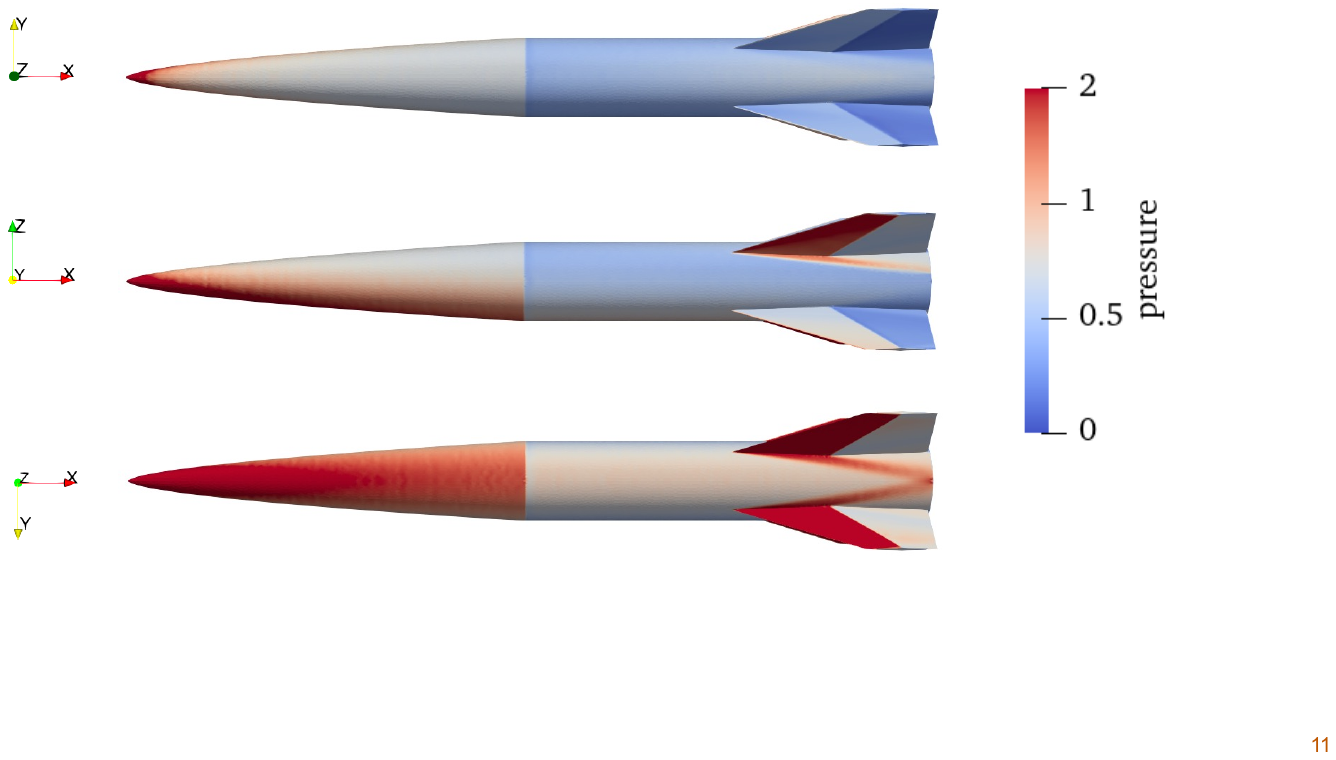}
	\caption{Top, side, and bottom view of surface pressure (non-dimensional) at flight conditions $M=6$, $\alpha=4$, and $\beta=0$. }
	\label{fig:ic3x_pressure}
\end{figure}

To demonstrate the MF linear regression methods, we consider the Initial Concept 3.X (IC3X) hypersonic vehicle. The IC3X was initially proposed by Pasiliao et al.~\cite{Pasiliao}, and a finite element model for the vehicle was developed by Witeof et al.~\cite{witeof2014initial}. A primary quantity of interest is the distributed aerodynamic pressure load over the surface of the vehicle at various flight condition parameters. Based on a nominal mission trajectory for this geometry, we consider the range of Mach numbers $M \in [5,7]$, angles of attack $\alpha \in [0,8]$, and sideslip angles $\beta \in [0,8]$. The surface pressure field is computed at a particular flight condition by solving the inviscid Euler equations using the flow solver package Cart3D~\cite{Cart3D, cart, aftosmis1997Cart3D} over an adaptive multilevel Cartesian mesh. The mesh adaptation scheme provides a natural hierarchy of model fidelity through various levels of mesh refinement. The discretized surface mesh remains constant, and contains $m=55966$ nodes, which is the dimension of the output surface pressure vector. An example non-dimensional surface pressure field solution computed by Cart3D at flight conditions of $M=6$, $\alpha=4$, and $\beta=0$ is visualized in Figure~\ref{fig:ic3x_pressure}. 

\subsection{Model specifications and data generation} \label{sec:fidelity}
We can construct different levels of fidelity for the pressure field solution by leveraging Cart3D's mesh adaptation, which refines the Cartesian volume mesh over multiple adaptation steps.  We define two levels of fidelity for simulating the surface pressure field: (i) the HF model with a finer volume mesh after more mesh adaptations and (ii) the LF model with a coarser volume mesh after fewer mesh adaptations and with a lower error tolerance. Specifically, we control the maximum number of initial mesh refinements (``Max Refinement"), the maximum number of adaptation processes (``Max Adaptations"), error tolerance, and the number of cycles per adaptation process (``Cycles/Adaptation") to generate the different fidelity levels. The specifications for the HF model and the LF model used in this work are described in Table~\ref{tab:fidelity_table}. We also provide the relative computational cost in terms of one HF model evaluation. Here, cost refers to the wall-clock time of running the HF and LF simulation on the same hardware. Note that we do not consider the LF simulations to be negligible cost and instead account for the cost of evaluating the LF samples when reporting the computational costs. 
\begin{table}[!htb]
    \caption{Model specifications \new{of hypersonic testbed problem}}\label{tab:fidelity_table}
        \centering
        \begin{tabularx}{\linewidth}{*{6}{>{\centering\arraybackslash}X}}
            \toprule
            \textbf{Model Type} & \textbf{Max Initial Refinement} & \textbf{Max\newline Adaptations} & \textbf{Error Tolerance} & \textbf{Cycles/\newline Adaptation} & \textbf{Cost Ratio} \\ \midrule
            \multirow{1}{*}{HF} & 7 & 12 & 1$\times$10$^{-3}$ & 175 & 1\\ \hline
            \multirow{1}{*}{LF} & 5 & 2 & 5$\times$10$^{-3}$ & 50 & 1/127\\ \bottomrule
        \end{tabularx} 
\end{table}

While the choice of HF and LF sample sizes is problem- and resource-dependent, in this case, we use a very limited number of HF samples $N_{\text{HF}} \in [3, 10]$, a LF training sample size of $N_{\text{LF}}=80$, and a HF testing sample size of $N^\text{test}_{\text{HF}}=50$ to analyze the effectiveness of the proposed methods in the ultra low-data regime. A large sample pool of 100 HF samples are drawn by Latin hypercube sampling (LHS). The testing set is then sampled via conditioned LHS \cite{minasny2006conditioned} from these points, and was fixed across all repetitions of the dataset. We bootstrap the remaining dataset by using conditioned LHS with different random seeds to create varying combinations of the training dataset and provide a measure of robustness of each method over 50 repetitions of the training samples (which entails the points for training are randomly distributed across the domain). We present the results while accounting for the computational cost of using the additional 80 LF samples given by $80/127=0.63$ equivalent HF samples.

\subsection{Results and discussion}\label{sec:re_mf}
\begin{figure}[h]
    \centering
    \includegraphics[width=\linewidth]{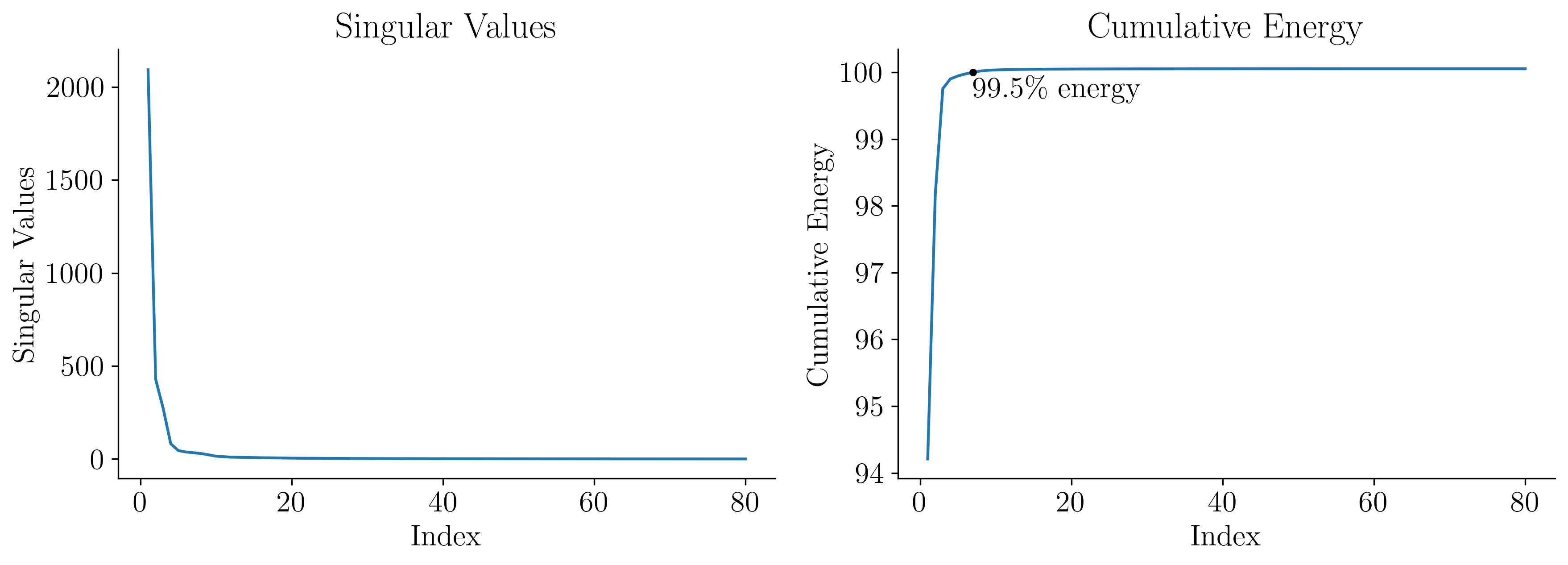}
    \caption{SVD on 50 repetitions of $N_{\text{LF}}=80$ LF training data\vspace{-1em}}
    \label{fig:LF_SVD_decay}
\end{figure}

\begin{figure}[h]
    \centering
    \includegraphics[width=\linewidth]{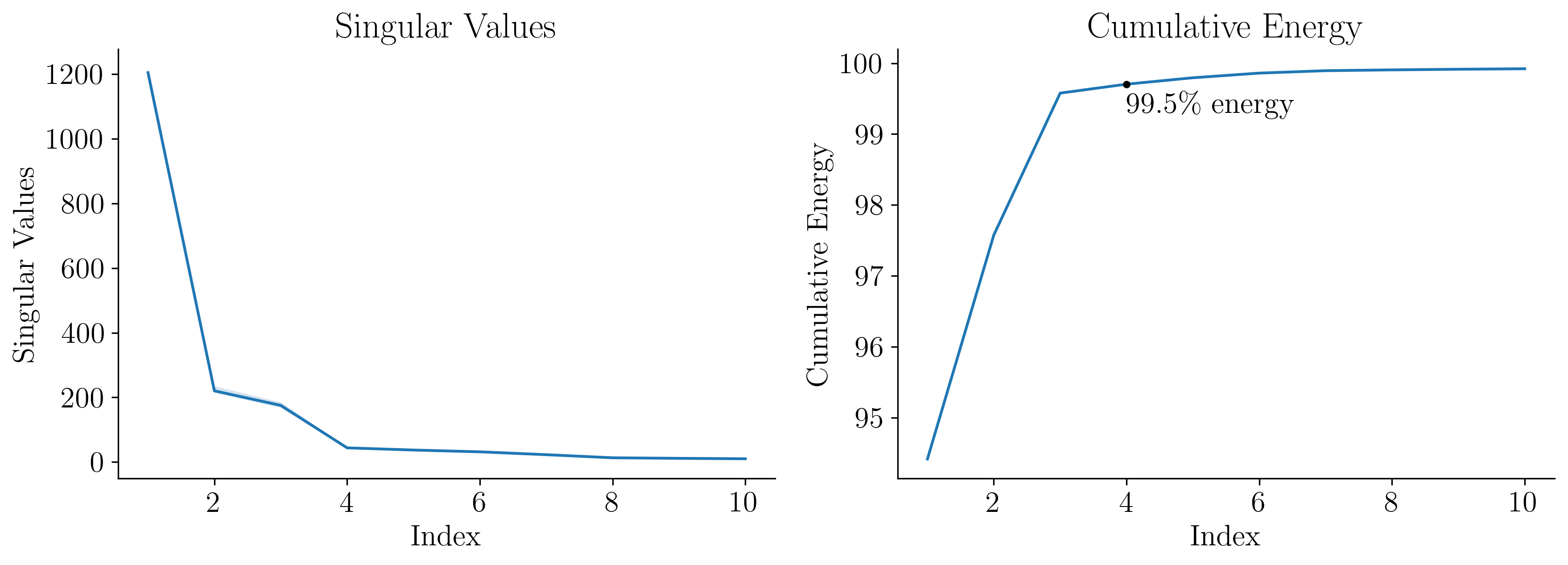}
    \caption{SVD on 50 repetitions of $N_{\text{HF}}=10$ HF training data\vspace{-1em}}
    \label{fig:HF_SVD_decay}
\end{figure}

We first analyze the dimensionality reduction on our training datasets of $N_{\text{HF}} = 10$ and $N_{\text{LF}} = 80$ to select an appropriate lower-dimensional subspace size. Figures~\ref{fig:LF_SVD_decay} and \ref{fig:HF_SVD_decay} show the singular value decay and the cumulative energy plots for the LF and HF data, respectively. We show the median of 50 repetitions of SVD computations and the 25th and 75th percentile shaded around the median curve. As is evident from the plots, there is not much variability in the singular values across the 50 different dataset draws. We use a tolerance of $\epsilon=0.995$ for the cumulative energy to decide the size of the low-dimensional subspace using \Eqref{eq:etol}. This leads to $k=7$ for most LF training datasets. For most HF training datasets with $N_\text{HF}>4$, we get $k=4$; otherwise, $k$ is bounded by number of HF training samples when $N_\text{HF}\le4$. This facilitates the use of lower dimensional representations of the data for the surrogate models to be trained on, without significant loss of information.

We apply the three MF linear regression methods described in Section \ref{sec:methods} to the prediction of the surface pressure field upon the IC3X testbed hypersonic vehicle. We evaluate the performance of a surrogate model through the normalized L2 accuracy metric given by $(1-\epsilon_{_\text{L2}})$, where the normalized L2 error $\epsilon_{_\text{L2}}$ is defined by %
\begin{equation}
    \epsilon_{_\text{L2}} \coloneq \frac{1}{N^\text{test}_\text{HF}} \sum_{i=1}^{N^\text{test}_\text{HF}} \frac{ \left\lVert \boldsymbol{y}^\text{HF}_i - \hat{\boldsymbol{y}}_i \right\rVert_{2}}{ \left\lVert \boldsymbol{y}_i \right\rVert_{2}}, \label{eq:l2_accuracy}
\end{equation} 
where $\lVert . \rVert_{2}$ is the L2 vector norm, $\boldsymbol{y}^\text{HF}_i $  is the HF model solution at $i^\text{th}$ test sample, and $\hat{\boldsymbol{y}}_i$ is the surrogate prediction at $i^\text{th}$ test sample. \new{The $R^2$ score is another generalizable and useful metric when comparing errors between high-dimensional quantities for regression methods~\cite{hastie2005elements}. We present the results using the $R^2$ metric in Appendix~\ref{sec:r2_results}.} Note that the results for the single-fidelity (SF) surrogate model refer to the linear regression which was trained on the HF pressure field data only. Since the surrogate models were trained on 50 varying repetitions of the training dataset, we present the median, 25th, and 75th percentiles of the test accuracies. For the SF model, the order of the polynomial was limited by the number of samples available -- limiting the choice to a linear equation in all cases. The MF linear regression with the additive structure also used a linear polynomial since it is trained on the same amount of HF data albeit with the discrepancy added. Lastly, both the MF surrogate models using the data augmentation methods were able to be trained using a polynomial of order two since the number of samples available to train was larger by the nature of the algorithms.

\begin{figure}[!htb]
\centering
\subfloat[Direct augmentation]{\label{fig:sample_weight_analysis}\includegraphics[width=.5\linewidth]{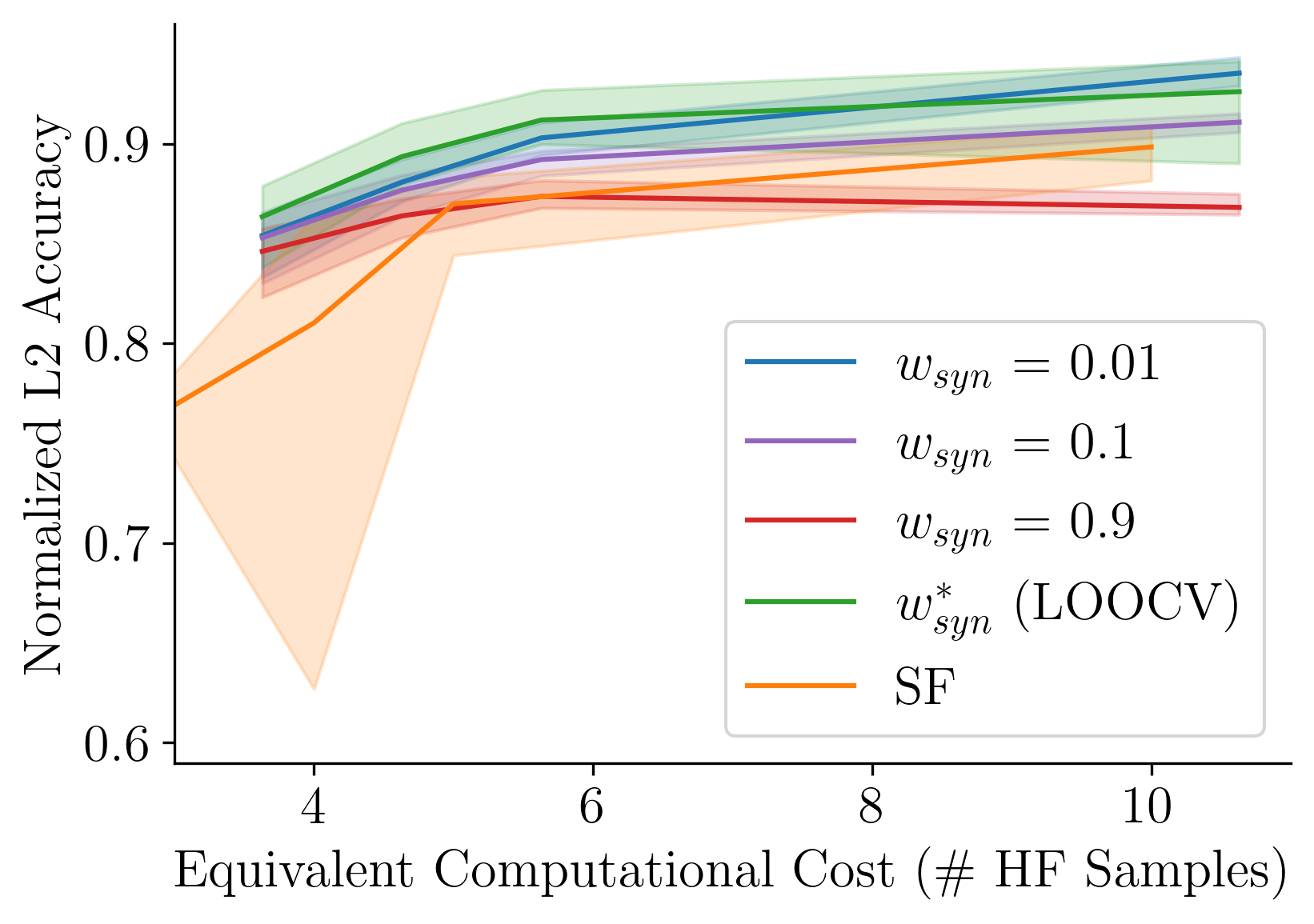}}\hfill
\subfloat[Explicit map]{\label{fig:sample_weight_analysis2}\includegraphics[width=.5\linewidth]{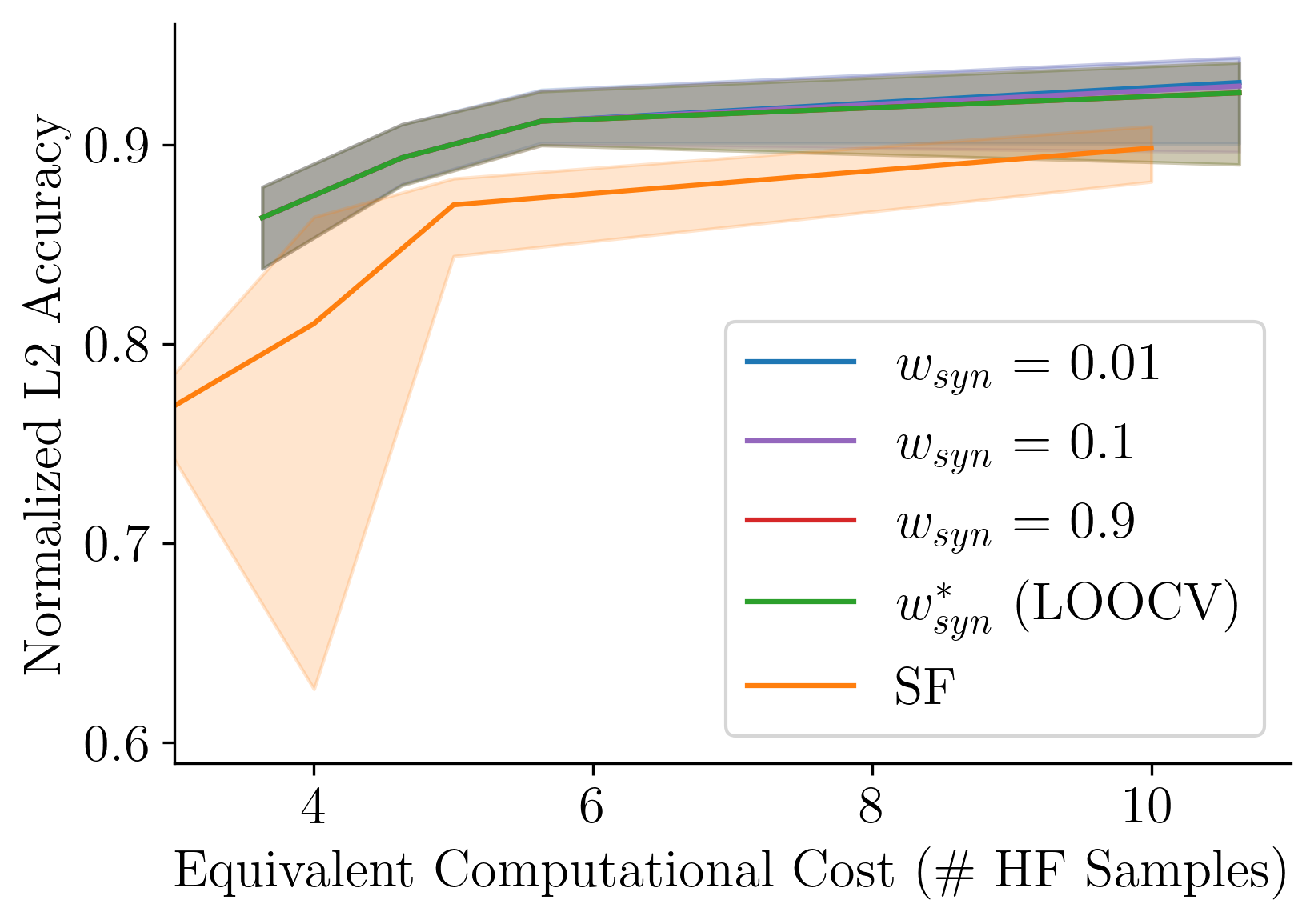}}
\caption{Comparison of weighting schemes for MF linear regression using data augmentation on 50 repetitions of the training dataset. Here, $w^*_{\text{syn}}$ refers to the optimal hyperparameter obtained after LOOCV for proximity-based weighting as described in \Eqref{eq:weighting_function}, and the other weights follow the fixed scheme.}
\label{fig:sample_weight}
\end{figure}

\begin{figure}[h]
\centering
\subfloat[Direct augmentation]{\label{fig:sample_weight_violin1}\includegraphics[width=.5\linewidth]{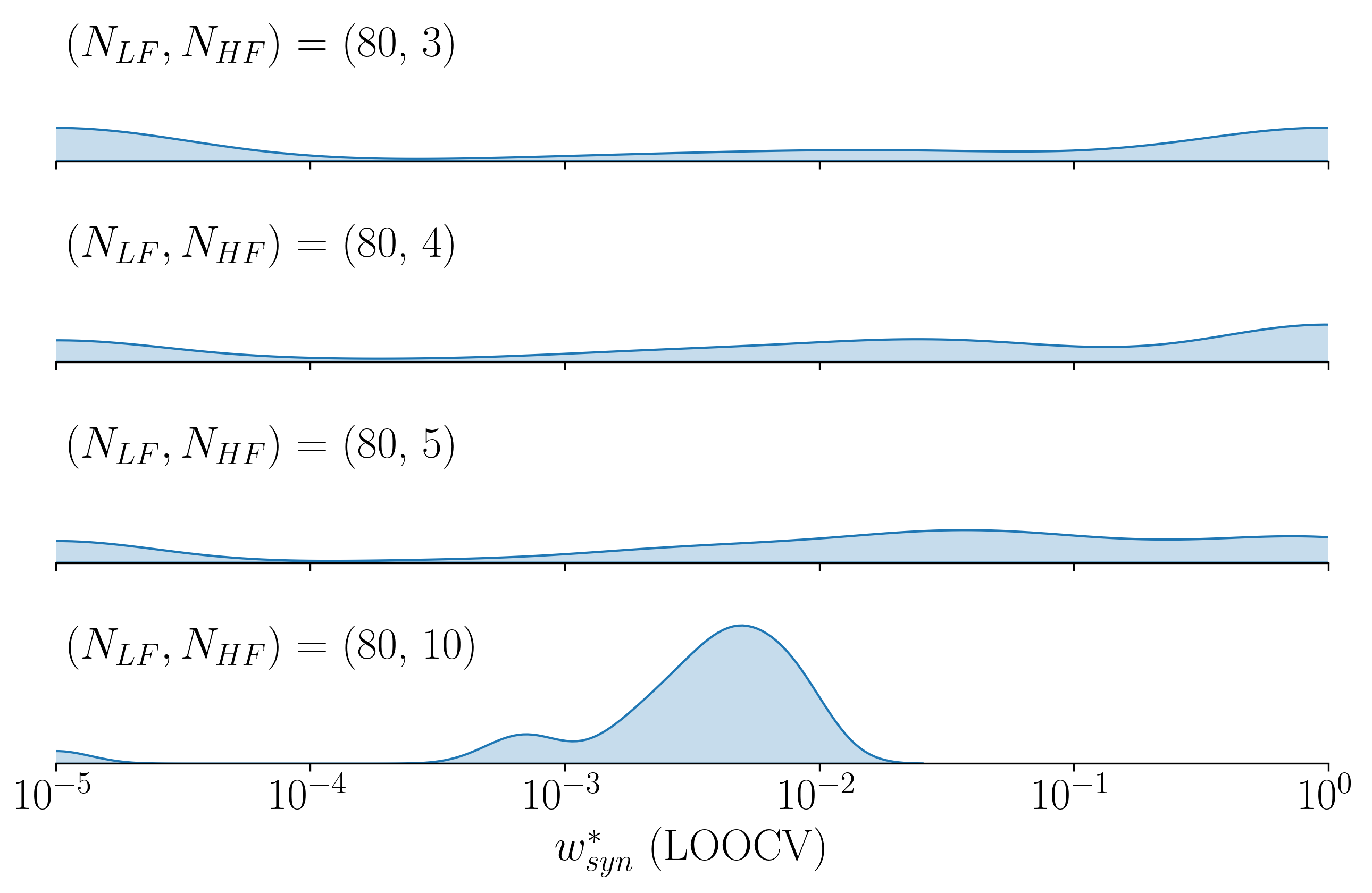}}\hfill
\subfloat[Explicit map]{\label{fig:sample_weight_violin2}\includegraphics[width=.5\linewidth]{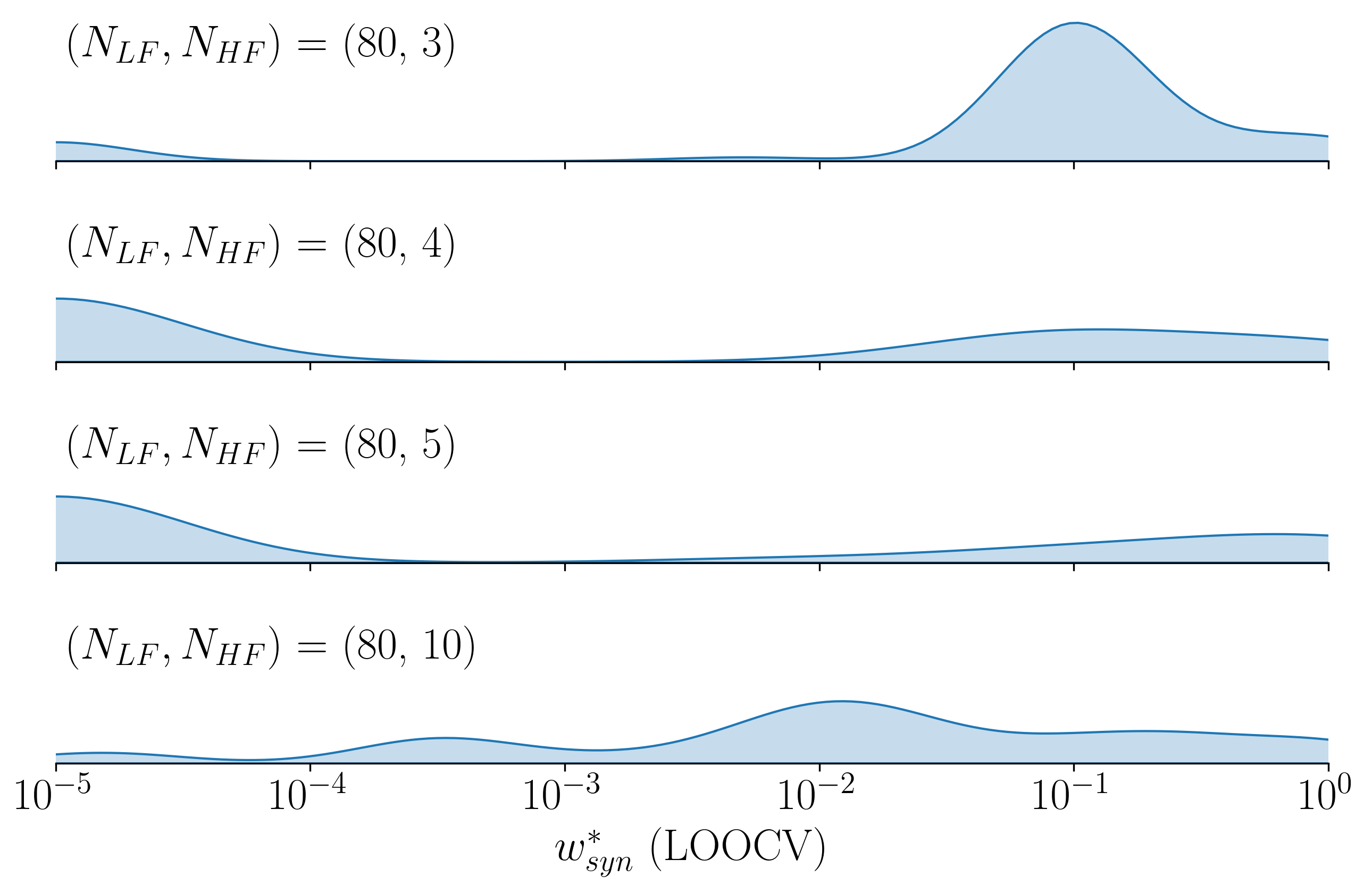}}
\caption{Comparison of $w^*_{\text{syn}}$ distributions obtained from LOOCV for MF linear regression using data augmentation with proximity-based weighting on 50 repetitions of the training dataset. Plotted are the kernel density estimates for each combination of LF and HF data.}
\label{fig:sample_weight_violin}
\end{figure}

Next, we analyze the impact of different sample weighting schemes on the results of the two data augmentation methods in Figure \ref{fig:sample_weight}. Setting the weights associated with the HF training samples to 1, we compare the fixed weighting scheme, where $w_{\text{syn}} \in \{0.01,0.1,0.9\}$ and $h(w_\text{syn})=w_{\text{syn}}$, against the LOOCV with proximity-based weighting method described in Eq~\eqref{eq:weighting_function}, where $h(w_\text{syn})=\sigma(\cdot ;w^*_{\text{syn}})$. Recall that $w^*_{\text{syn}}$ is the optimal weighting function hyperparameter value for proximity-based weighting obtained through the LOOCV procedure described in Section~\ref{sec:c-val}. For this application we use the Heaviside step function to implement $\sigma(\cdot ;w^*_{\text{syn}})$, with a threshold $\tau$ set to eliminate the furthest 10th percentile of LF samples (by minimum Euclidean distance to HF samples). We observe that the direct data augmentation method is sensitive to the choice of $w_\text{syn}$ for the fixed weighting scheme, with a variation of up to $\sim10$\% in median accuracy. On the contrary, the explicit map data augmentation method is less sensitive to changes in the sample weight, with a variation of up to 2\% in median accuracy. We find that the LOOCV method for determining $w^*_\text{syn}$ for each repetition of the training dataset performs close to the best fixed weighting scheme option for both data augmentation methods. This highlights the effectiveness of automatic weight selection based on the underlying training dataset. Figure \ref{fig:sample_weight_violin} shows the distribution of optimized LF sample weights across 50 repetitions of HF and LF training datasets, following the LOOCV-based optimization procedure. The plotted quantity corresponds to the value of $w^*_{\text{syn}}$ obtained through LOOCV for proximity-based weighting function described in \Eqref{eq:weighting_function} (here, $w^*_{\text{syn}}$ is the maximum weight possible when using the Heaviside function).
The distribution of $w^*_\text{syn}$ across the 50 training repetitions is generally bimodal in our setting with $N_\text{HF}\le 10$. This bimodality arises in part because the LOOCV optimization is initialized at $10^{-1}$, which explains the presence of a higher mode near this magnitude. As the HF sample size increases, the resulting weight distribution shifts toward smaller magnitudes, suggesting that the added HF data reduces the reliance on LF information for accurate prediction for this application.


\begin{figure}[!h]
\centering
\subfloat[Comparison of all methods]{\label{fig:mfcomparea}\includegraphics[width=.5\linewidth]{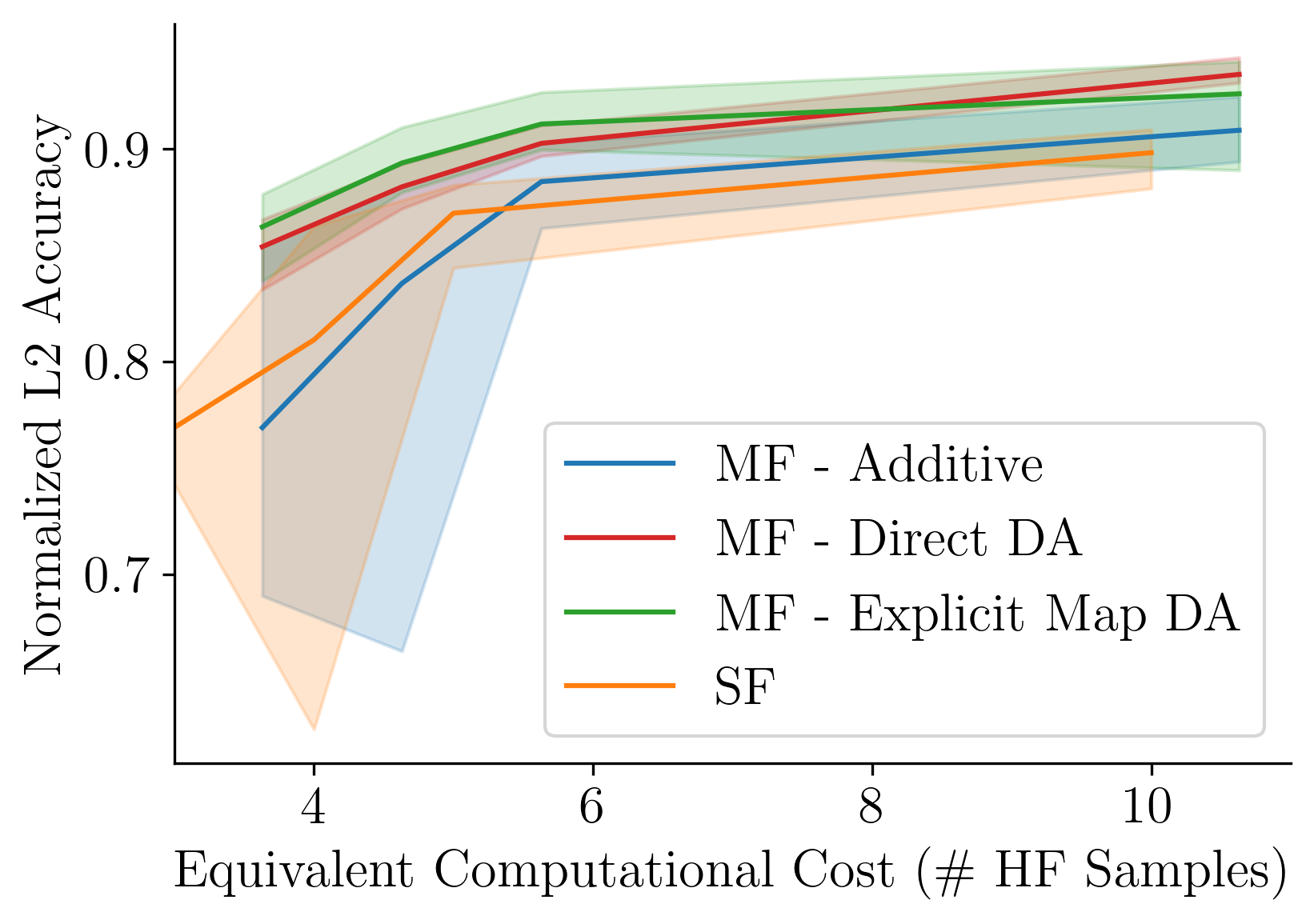}}\hfill
\subfloat[Additive method]{\label{mfcompareb}\includegraphics[width=.5\linewidth]{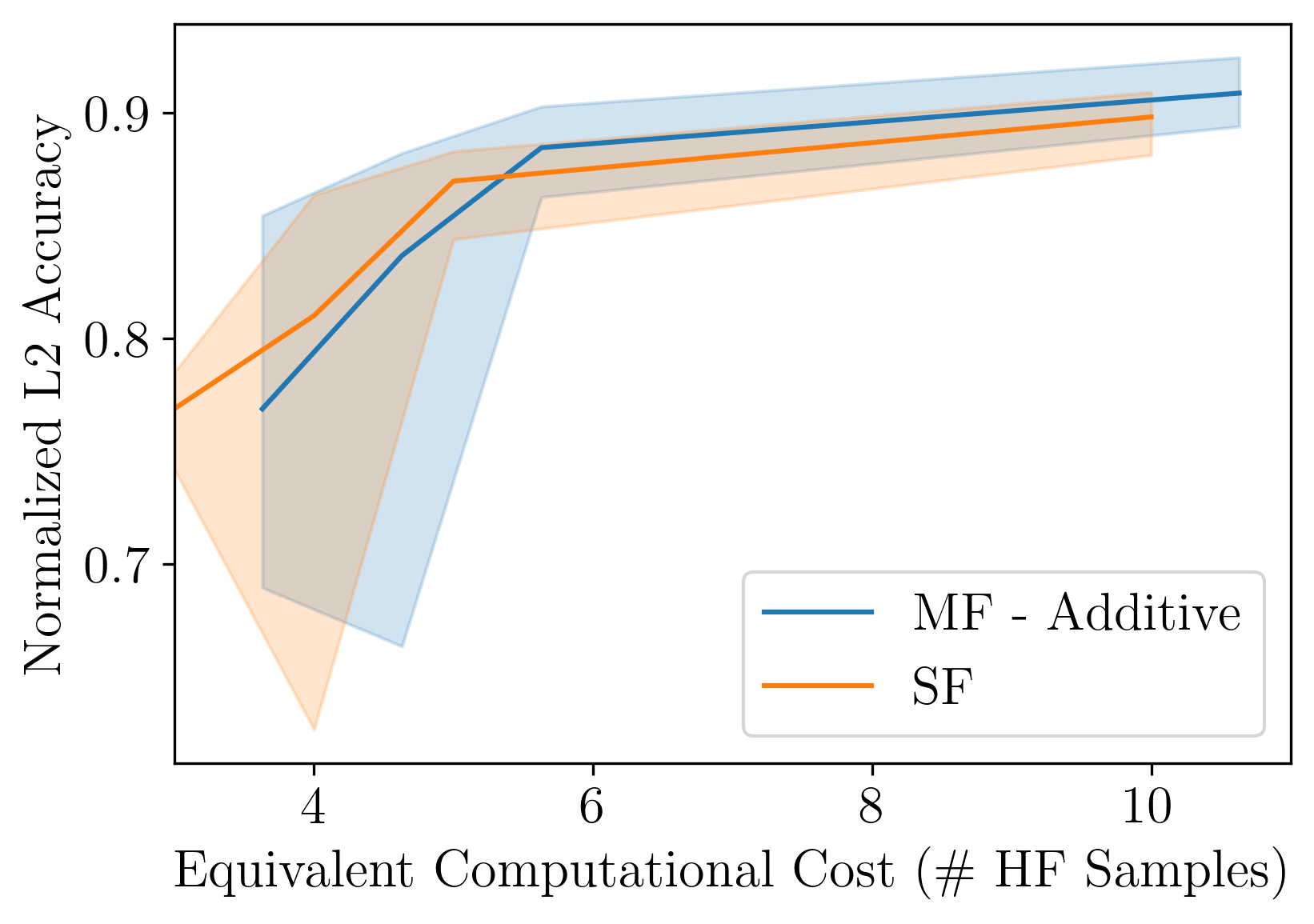}}\par 
\subfloat[Data augmentation: direct augmentation]{\label{mfcomparec}\includegraphics[width=.5\linewidth]{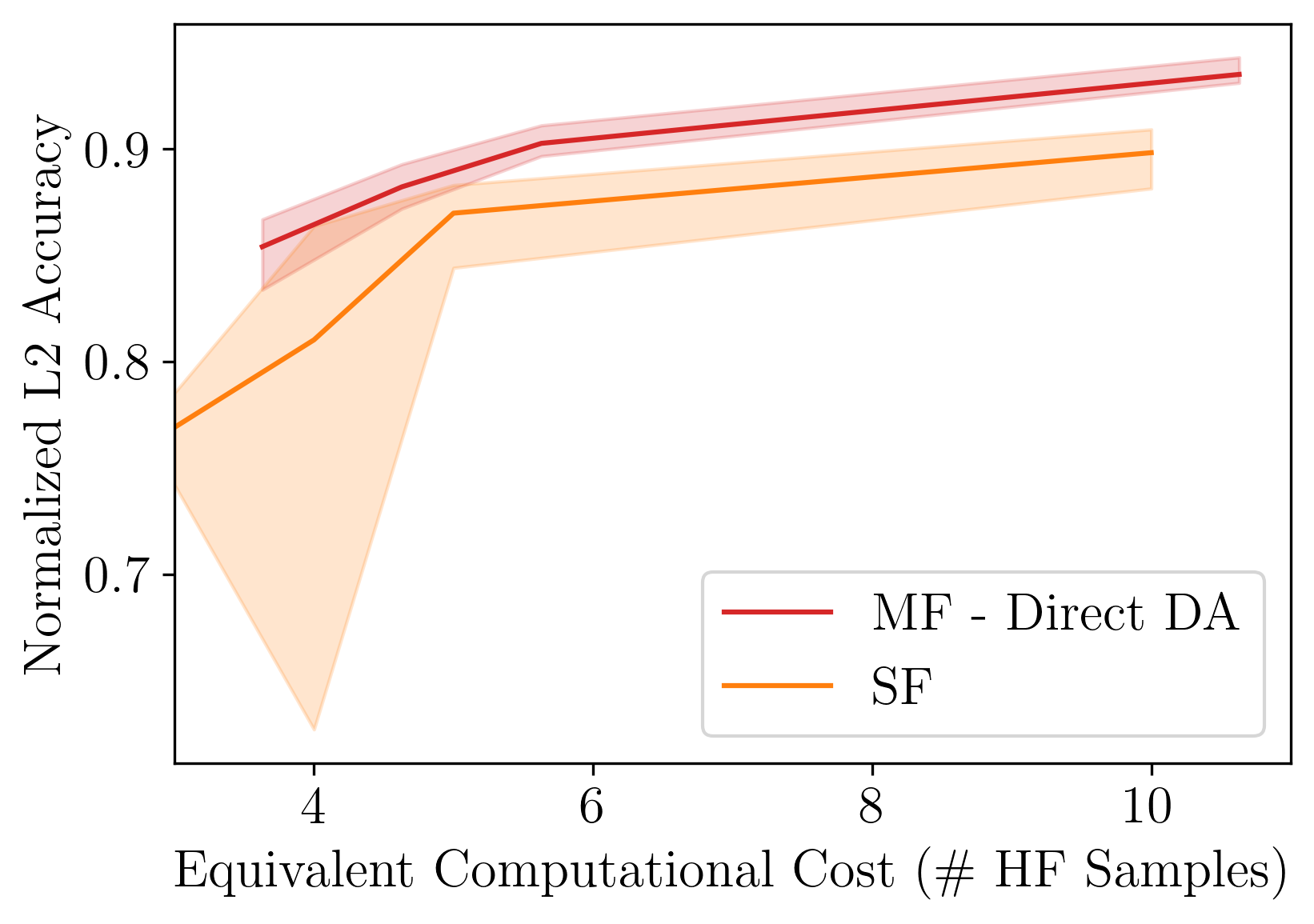}}\hfill
\subfloat[Data augmentation: explicit mapping]{\label{mfcompared}\includegraphics[width=.5\linewidth]{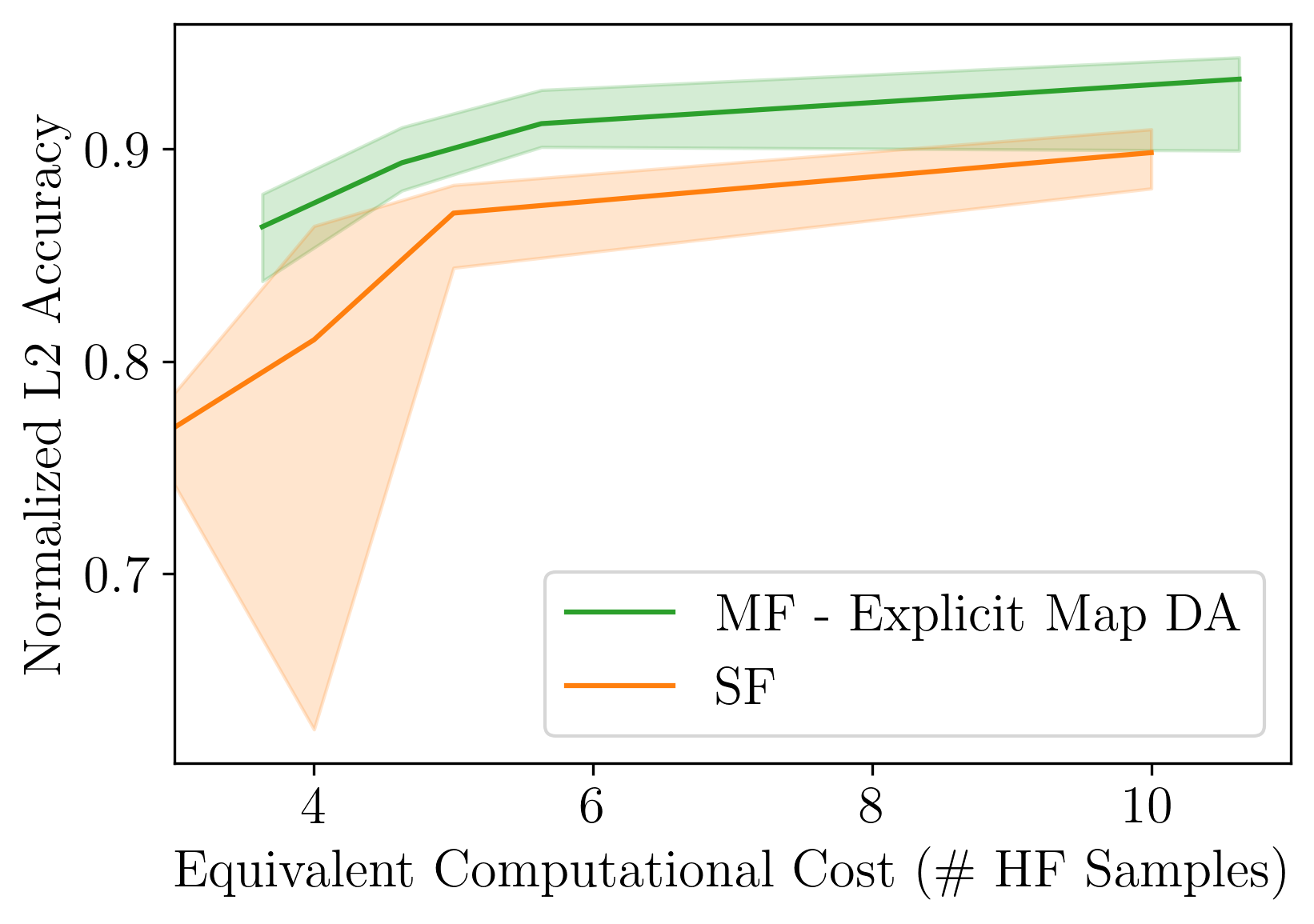}}
\caption{\new{Comparison of proposed MF linear regression methods to baseline SF linear regression and the additive MF method} on 50 repetitions of the training dataset (DA denotes a data augmentation method implemented with LOOCV and the proximity-based weighting)}
\label{fig:mfcompare}
\end{figure}

Figure \ref{fig:mfcompare} \new{shows the comparison of the two MF linear regression methods proposed in this work and the additive MF method presented in Appendix~\ref{app:mf-add}, in comparison with the SF surrogate model.} The additive MF method performs similar to the SF linear regression and does not offer significant increase in accuracy for this application. 
In contrast, both the data augmentation techniques (using the optimal $w_\text{syn}$ after LOOCV and proximity-based weights) perform better than the additive approach and show a substantial improvement in accuracy over the SF linear regression for equivalent computational cost. Furthermore, the robustness of both the MF linear regression models with data augmentation is markedly better than the SF surrogate model. This is likely due to the fact that the MF linear regression model sees a larger variety of data during the training phase. The extra LF samples in the data augmentation methods are of course not fully representative of the HF model, as indicated by sample weights of 1 for the HF samples and $w^*_\text{syn}<1$ for the synthetic data generated from the LF samples as seen in Figure~\ref{fig:sample_weight_violin}. The MF method with explicit map for data augmentation performs the best with few samples, while the direct augmentation had the highest accuracy with the largest amount of training data. Table~\ref{tab:KOH_CART_table} provides the median test accuracies \new{and $R^2$ score} of each regression method for $N_\text{HF}=$ 3, 5, and 10 HF samples. We find that the data augmentation technique using explicit map leads to an improvement of approximately $9.5\%$ compared to the SF model for $N_\text{HF}=3$ HF samples and $3.2\%$ compared to the SF model for $N_\text{HF}=10$ HF samples. Interpolating at the first sample size tested for the MF methods, $N_\text{HF}= 3.63$ (3 HF, 80 LF samples), yields a 12.4\% improvement in accuracy compared to the SF method. \new{With respect to the $R^2$ values, the results of the data augmentation methods were shown to have similar improvements upon the SF and additive-based approaches.}
\begin{table}[!h]
\caption{Selected multifidelity linear regression results \new{(DA denotes a data augmentation method implemented with LOOCV and the proximity-based weighting)}}
    \centering
    \begin{tabularx}{\linewidth}{l*{4}{>{\centering\arraybackslash}X}}
        \toprule
        \textbf{Model Type} & \textbf{\# LF Samples} & \textbf{\# HF Samples} & \textbf{Median Normalized L2 Test Accuracy} & \textbf{\new{Median Test Data $R^2$ Score}}\\ \midrule
        \multirow{3}{*}{SF} & - & 3 & 0.768  & {0.38}\\ 
         & - & 5 & 0.870  & {0.85} \\ 
         & - & 10 & 0.898 & {0.90} \\ \hline
        \multirow{3}{*}{MF - Additive} & \multirow{3}{*}{80} & 3 & 0.763 & {0.52} \\ 
         &  & 5 & 0.875 & {0.88} \\
         &  & 10 & 0.909 & {0.91} \\ \hline
         \multirow{3}{*}{\shortstack{MF - Direct DA (LOOCV $w^*_\text{syn}$)}} & \multirow{3}{*}{80} & 3 & 0.854 & {0.70}\\ 
         &  & 5 & 0.903 & {0.86}\\
         &  & 10 & 0.935 & {0.94}\\ \hline
         \multirow{3}{*}{\shortstack{MF - Explicit map DA (LOOCV $w^*_\text{syn})$}} & \multirow{3}{*}{80} & 3 & 0.863 & {0.80}\\ 
         &  & 5 & 0.912 & {0.93}\\
         &  & 10 & 0.930 & {0.95}\\ 
         \bottomrule
    \end{tabularx}
    \label{tab:KOH_CART_table}
\end{table}

\begin{figure}[!htb]
	\centering
	\includegraphics[width=\textwidth]{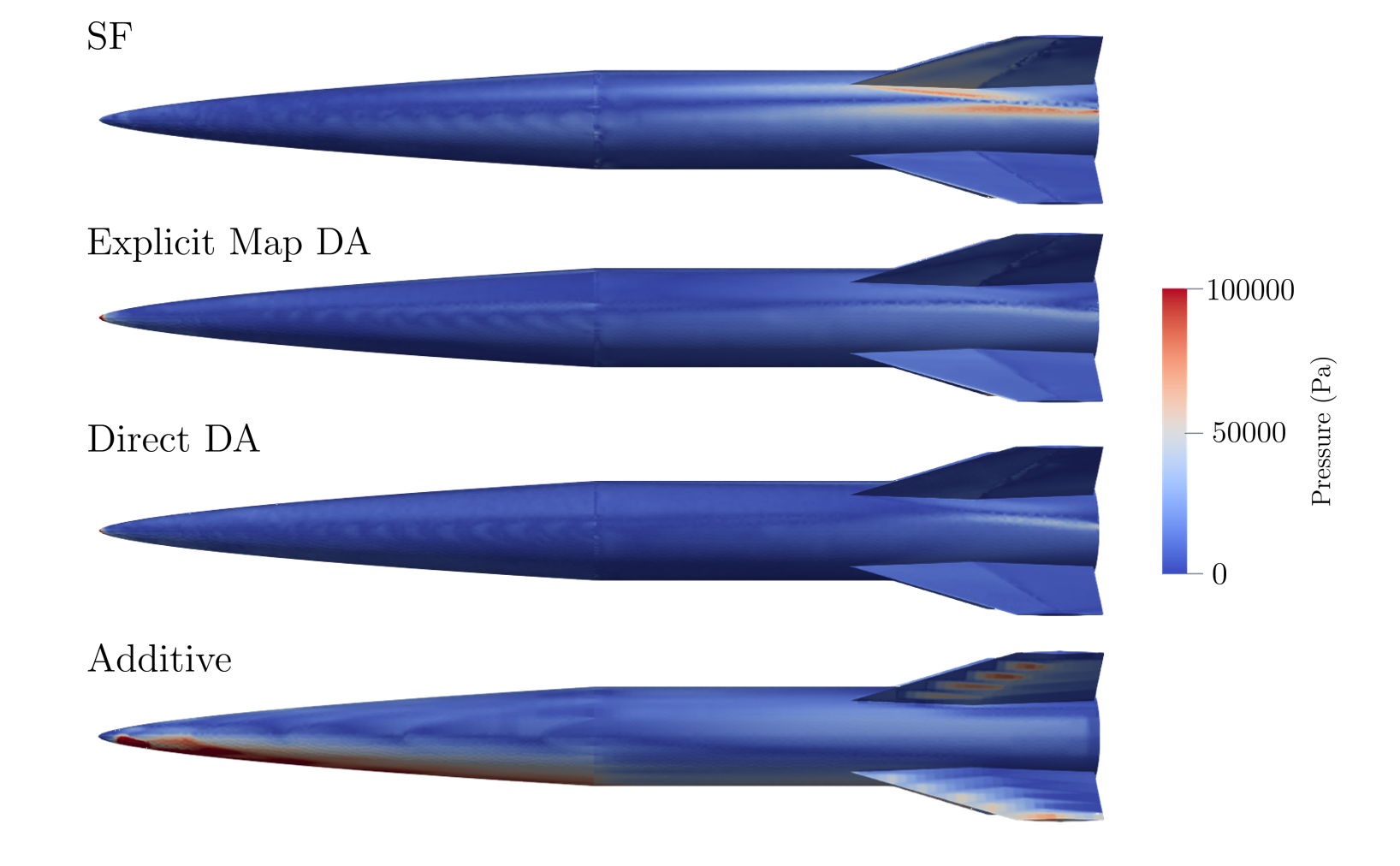}
	\caption{Comparison of errors in pressure field prediction at \textbf{Mach 6.79,} $\boldsymbol{\alpha = 4.97^{\circ}, \beta = 4.74^{\circ}}$}
	\label{fig:two_rockets}
\end{figure}

Finally, we look at a comparison of the absolute errors in pressure prediction between the SF surrogate and the MF surrogate methods. For an arbitrary test sample, we predict the pressure field using the surrogates and show the absolute error compared to the HF model simulation. We show a contour plot of the errors on the vehicle body in Figure \ref{fig:two_rockets}, providing some visual context for the gains the MF surrogate model nets.

\section{\new{Numerical demonstration: aircraft disc brakes}}
\label{sec:num_example2}
\new{
In this section, we present the results for an aircraft disc brake thermal modeling problem described in Section~\ref{sec:brakes}. The HF and the LF models used for the MF linear regression are described in Section~\ref{sec:fidelity2}. Then, we present results for the projection-based MF linear regression methods proposed in this work in Section~\ref{sec:re_mf2} and show the comparison to the state-of-the-art additive approach for MF regression.
}

\begin{figure}[!htb]
\centering
\subfloat[\new{Brakes off}]{\label{fig:no_braking}\includegraphics[width=.5\linewidth]{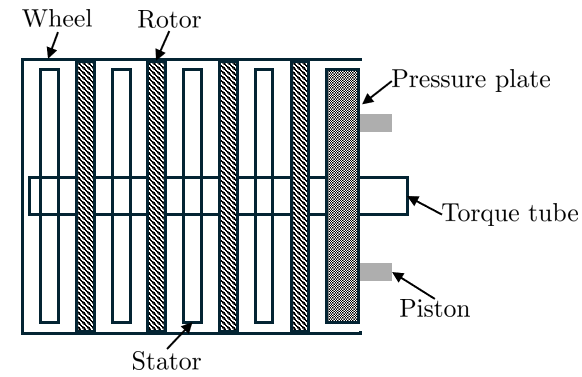}}\hfill
\subfloat[\new{Brakes on}]{\label{fig:under_braking}\includegraphics[width=.5\linewidth]{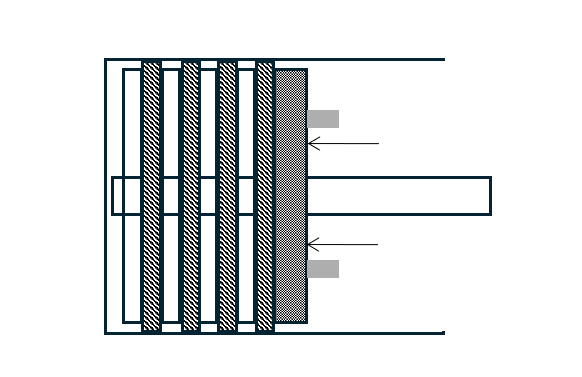}}
\caption{\new{Notional illustration of multi-disc transport-aircraft brakes, adapted from~\cite{daidzic2017modeling}}}
\label{fig:aircraft_brakes}
\end{figure}

{
\subsection{Aircraft disc brake thermal problem description}\label{sec:brakes}
}
\new{Aircraft wheel braking systems are designed to be capable of ensuring safe deceleration and stopping performance following landing or during rejected takeoff (RTO) scenarios, where air brakes are less impactful. During braking, the kinetic energy of the aircraft is rapidly converted into thermal energy through friction at the brake rotors and stators in a multi-disc brake setup, leading to substantial temperature rises within the brake stack. For aircraft such as the Boeing 737 (B737), each of the four main landing gear wheels is equipped with four rotor–stator pairs, which collectively absorb tens of megajoules of kinetic energy during a typical stop~\cite{Brady_2015}. 
When braking pressure is applied, multi-disc aircraft brakes dissipate kinetic energy through thermal energy by clamping each of the rotor-stator pairs together through a hydraulic piston.
A notional drawing of aircraft brakes in a typical transport-aircraft is depicted in Figure~\ref{fig:aircraft_brakes}. Understanding the temperature evolution and spatial distribution within the brake discs is essential for assessing brake deterioration and the turnaround time required before the next takeoff. Additionally, efficient modeling of the temperature field given braking duration, aircraft mass, approach velocity, and other conditions can be utilized to expedite optimization and design cycles.}

\new{We develop an axisymmetric finite element model to simulate the transient temperature field within the brake stack of a representative narrow-body transport aircraft, based on the B737 geometry and braking configuration. The model resolves the temperature over an annular rotor–stator stack during a braking event of duration $\tau_b$, by representing the frictional heating as a heat source term. Furthermore, convective cooling along all the exposed surfaces is modeled. For the MF linear regression, the primary quantity of interest is the temperature field $T(r,z)$ at the end of the braking period where $r$ and $z$ are the radial and axial coordinates in meters. One potential application of such a surrogate model is to rapidly evaluate designs by checking where maximum temperature occurs and if maximum allowable temperature limits are met in the brake assembly given design changes. We consider six model inputs for the surrogate model that include physical and environmental parameters in the form of aircraft mass, approach velocity (or ground velocity in the case of an RTO), braking duration, material thermal diffusivity, ambient temperature, and initial temperature. Table \ref{tab:input_table} summarizes the parameter definitions and ranges used in this example which were chosen to reflect typical flight conditions of the B737 aircraft.}
{
\begin{table}[!htb]
    \caption{\new{Input parameters and their ranges for the aircraft brake temperature field problem}}\label{tab:input_table}
        \centering
        \begin{tabularx}{0.8\linewidth}{>{\arraybackslash}X>{\centering\arraybackslash}X}
            \toprule
            \textbf{Input Parameter (Units)} & \textbf{Range} \\ \midrule
            Aircraft mass, $m$ (kg) & [67720, 85140] \\
            Landing velocity, $V$ (m/s) & [30.0, 75.0] \\
            Braking duration, $\tau_b$ (s) & [10, 30] \\
            Thermal diffusivity, $\alpha$ (m$^2$/s) & [6$\times$10$^{-6}$, 2$\times$10$^{-5}$] \\
            Ambient temperature, $T_{\infty}$ (K) & [253.15, 319.15] \\
            Initial disc temperature, $T_{\mathrm{initial}}$ (K) & [373.15, 473.15] \\ 
            \bottomrule
        \end{tabularx}
\end{table}
}

{
\subsection{Model specifications and data generation} \label{sec:fidelity2}
}
\new{
The thermal analysis of the aircraft disc brakes involves solving the 2D transient axisymmetric heat equation represented by the governing equation
\begin{equation}
    \begin{split}
    \rho c_p \frac{\partial T}{\partial t} - \nabla \cdot (k \nabla T) &= f(r,z,t),\\ 
    T(r,z,0) &= T_{\mathrm{initial}},
\end{split}
\label{eq:heat_eqn}
\end{equation}
where $T(r,z,t)$ K is the temperature field at time $t$, $\rho$ kg/m$^3$ is the density, $c_p$ J/kg$\cdot$K is the isobaric specific heat capacity, $k$ W/m$\cdot$K is the thermal conductivity, and $f(\cdot)$ W/m$^3$ is the volumetric heat source term which represents the frictional heating at the rotor-stator interfaces. The initial condition represents the disc brakes having an initial uniform temperature $T_{\mathrm{initial}}$. The boundary conditions representing the convective cooling through airflow around the exterior surfaces and an adiabatic boundary at the axle or torque tube connection is given by
\begin{equation}
    \begin{split}   
    -k \nabla T \cdot n &= h(T-T_\infty) \quad \text{on outer radius and end-plates} \\
    -k \nabla T \cdot n &= 0 \quad \text{on inner radius},
\end{split}
\label{eq:heat_eqn_bc}
\end{equation}
where $h$ is the convective heat transfer coefficient of air. }

\new{The frictional heating source term is given by the following expression,
\begin{align}f(r,z,t) &=
\begin{cases}
\displaystyle\sum_{k=1}^{N-1} \dfrac{q_k(t)}{\varepsilon} \mathbf{1}_{k}(r,z), & 0 \leq t \leq \tau_b \\
0, & t > \tau_b
\end{cases}
\label{eq:heat_source}
\\
\mathbf{1}_{k}(r,z) &=
\begin{cases}
1, & |z-z_k| \leq \tfrac{\varepsilon}{2} \\
0, & \text{otherwise}.
\end{cases}
\label{eq:indicator_func}
\end{align}
where $q_k(t)$ (W/m$^2$) is the heat flux as a function of time at the $k$-th interface between a rotor-stator pair and $\varepsilon$ (m) is chosen to be sufficiently large enough to encapsulate enough elements at the interface locations. Given $N$ rotor-stators, frictional heating will occur at the $N-1$ interfaces between them. We are assuming that perfect contact occurs between all rotor-stator pairs without resistance and the rotor-stators are of equal size as in~\cite{ramadan2021numerical, meunier2022thermal}; hence, we sum the individual heating contributions equally. In \Eqref{eq:indicator_func}, an indicator function $\mathbf{1}_{k}(r,z)$ is used to ensure heating only occurs at the small epsilon-sized band centered around each rotor-stator contact interface. We show in Appendix \ref{app:conserve_source} that this formulation correctly conserves the braking energy the brakes must dissipate.
}

\new{
The governing equation in \Eqref{eq:heat_eqn} and the boundary conditions are used to construct the variational form of the problem, which is passed to the FEniCSx platform~\cite{BarattaEtal2023,ScroggsEtal2022,BasixJoss,AlnaesEtal2014} to solve numerically. We use quadrilateral mesh elements and the backward Euler method for discretization in time. To define two levels of fidelity for this problem, we set up the HF model with the finest discretization in time and space and the LF model with a far coarser discretization. The exact specifications for the HF and LF model are detailed in Table~\ref{tab:fidelity_table2}. As a result of the spatial discretization there are $m = $ 51681 nodes in the HF simulation. We again consider a data-sparse environment with limited HF samples, specifically $N_\text{HF} \in [4,12]$ and a LF training size of $N_\text{LF}=25$. The same procedure using bootstrapping as described in Section~\ref{sec:fidelity} is used to generate the results and provide a measure of robustness, wherein a pool of 150 HF samples are drawn and then sampled from. The HF testing set is fixed across all training dataset repetitions and is of size $N_\text{HF}^\text{test}=100$.}
\begin{table}[!htb]
    \caption{\new{Model specifications of aircraft brake problem}}\label{tab:fidelity_table2}
        \centering
        \begin{tabularx}{\linewidth}{*{5}{>{\centering\arraybackslash}X}}
            \toprule
            \textbf{Model Type} & \textbf{$n_r$ (radial elements)} & \textbf{$n_z$ (axial elements)} & \textbf{$\Delta$t (s)} & \textbf{Cost Ratio} \\ \midrule
            \multirow{1}{*}{HF} & 320 & 160 & 2.5$\times$10$^{-2}$ & 1 \\ \hline
            \multirow{1}{*}{LF} & 40 & 20 & 5$\times$10$^{-1}$ & 1/344 \\ \bottomrule
        \end{tabularx} 
\end{table}

\subsection{\new{Results and discussion}}\label{sec:re_mf2}
\begin{figure}[!htb]
    \centering
    \includegraphics[width=\linewidth]{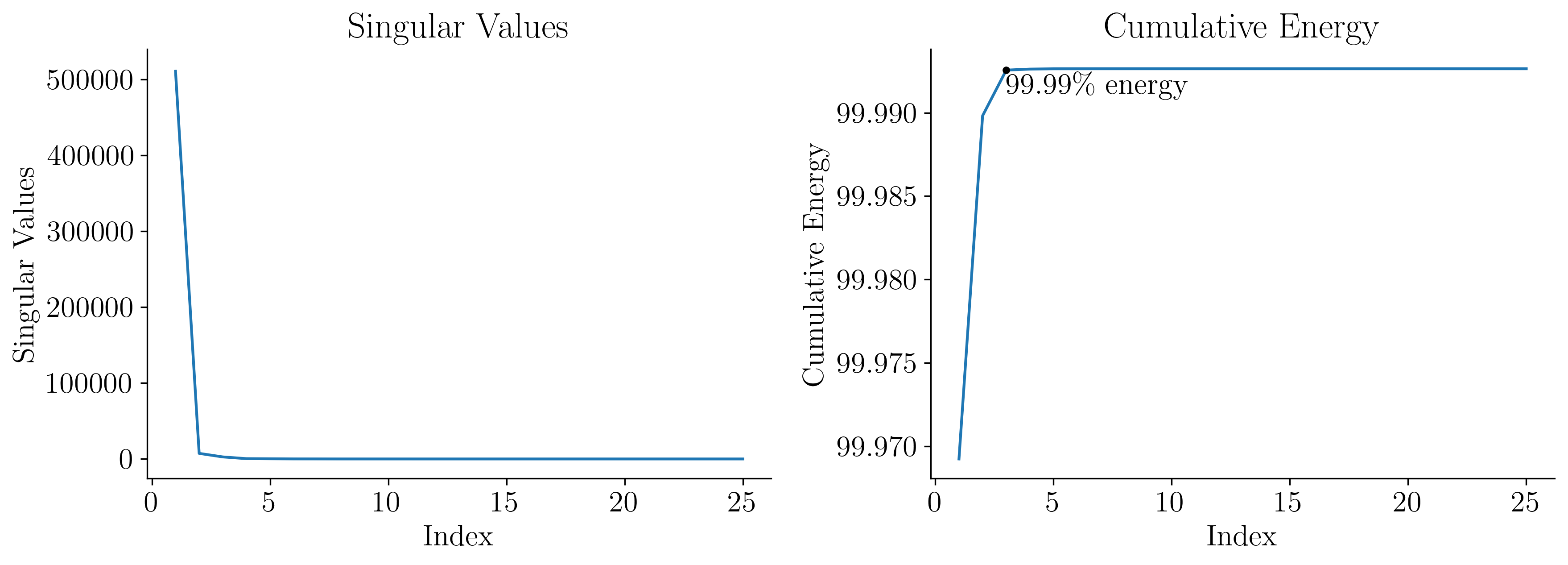}
    \caption{SVD on 50 repetitions of $N_{\text{LF}}=25$ LF training data\vspace{-1em}}
    \label{fig:LF_SVD_decay_brake}
\end{figure}

\begin{figure}[!htb]
    \centering
    \includegraphics[width=\linewidth]{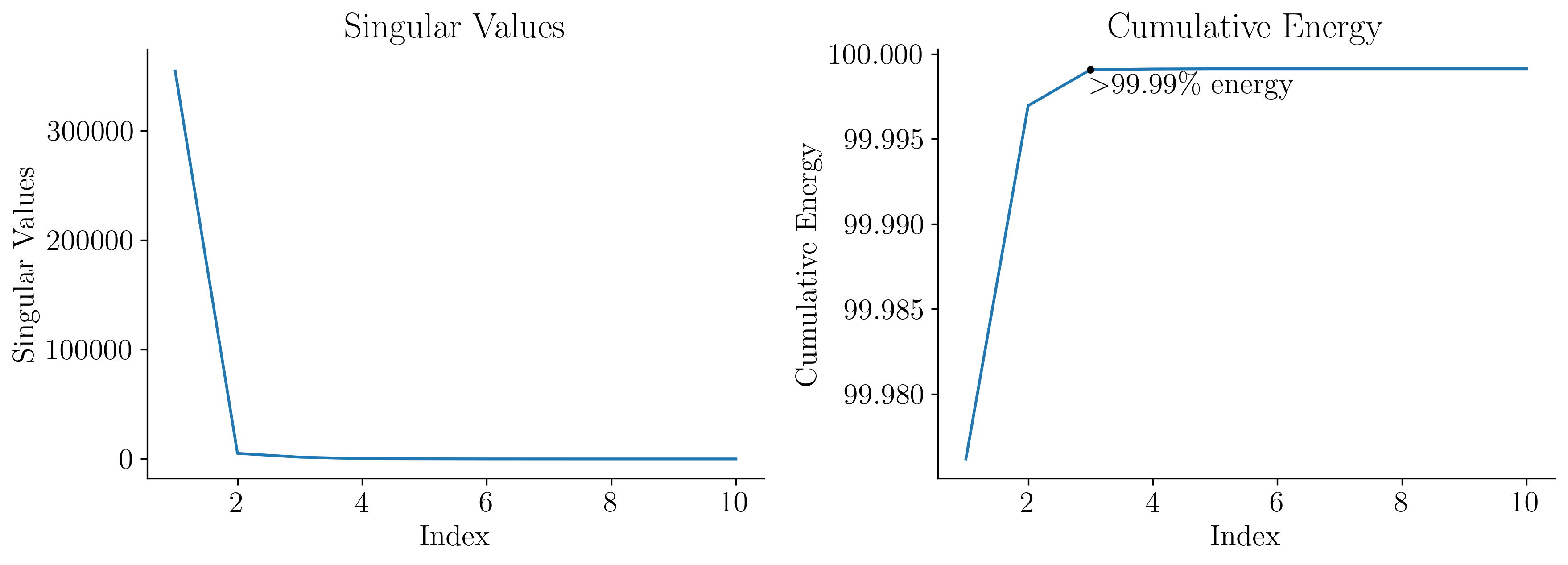}
    \caption{SVD on 50 repetitions of $N_{\text{HF}}=12$ HF training data\vspace{-1em}}
    \label{fig:HF_SVD_decay_brake}
\end{figure}

\begin{figure}[!htb]
\centering
\subfloat[Direct augmentation]{\label{fig:sample_weight_analysis_brake}\includegraphics[width=.5\linewidth]{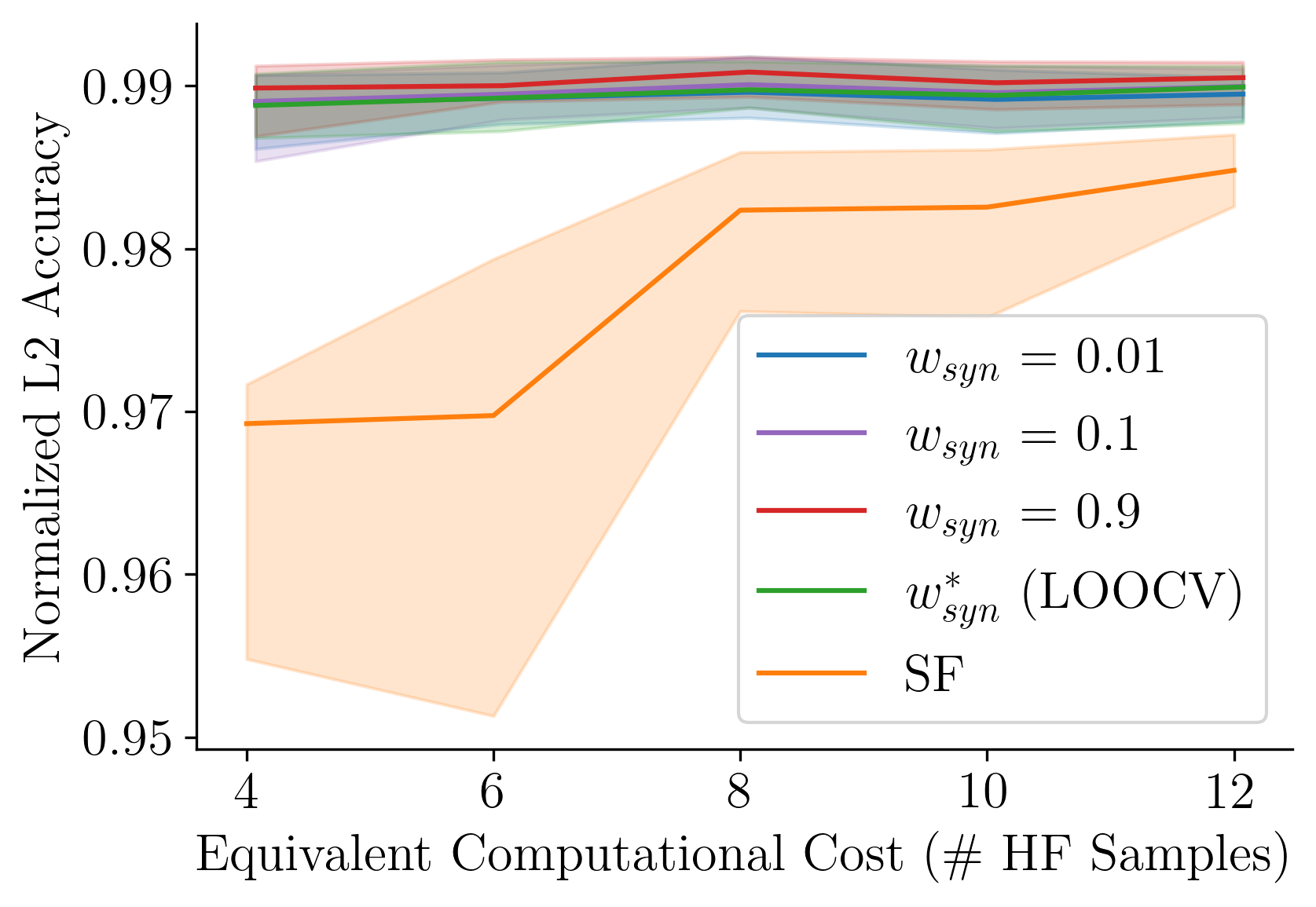}}\hfill
\subfloat[Explicit map]{\label{fig:sample_weight_analysis2_brake}\includegraphics[width=.5\linewidth]{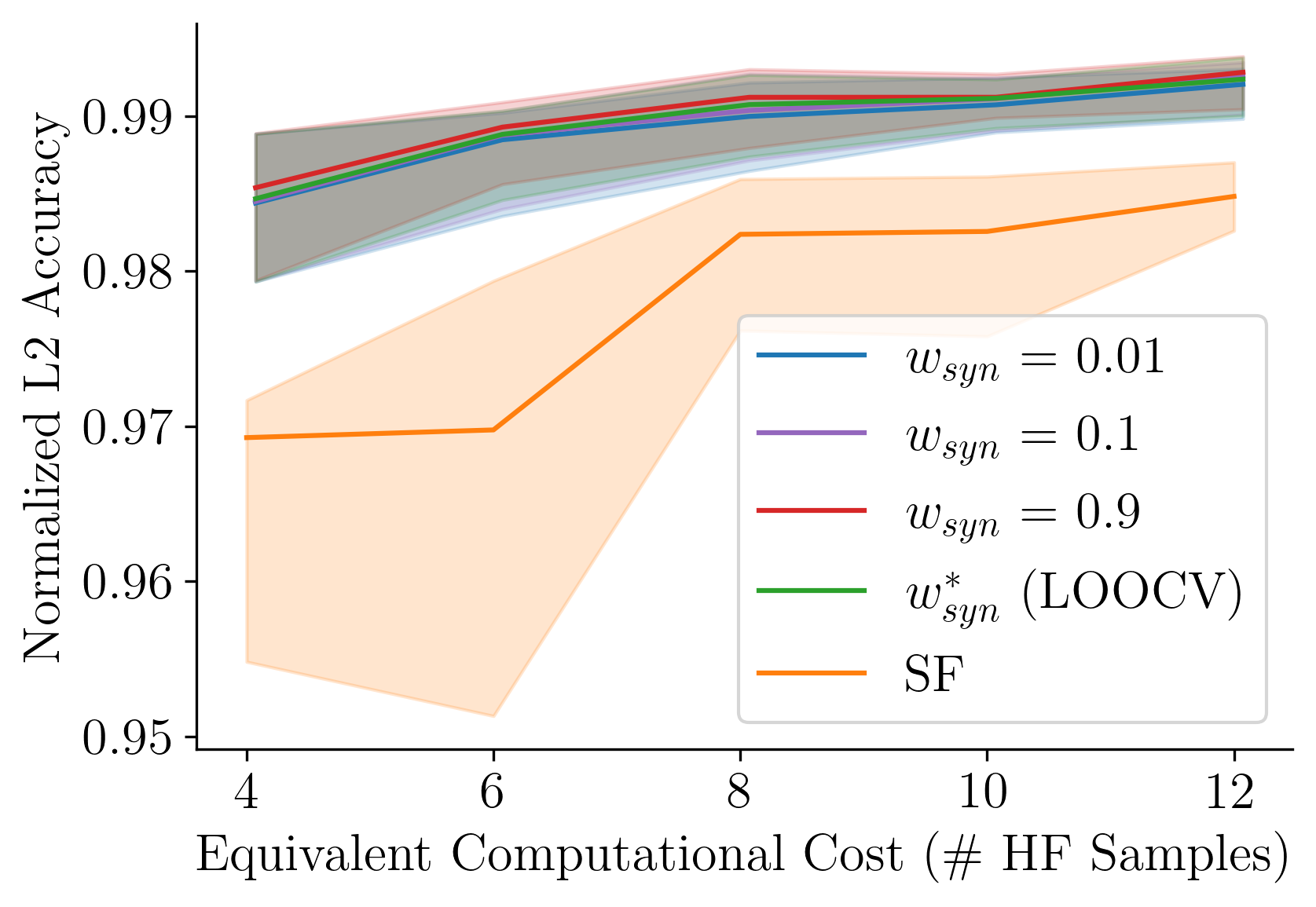}}
\caption{\new{Comparison of weighting schemes for MF linear regression using data augmentation on 50 repetitions of the training dataset for the aircraft brakes thermal modeling problem. Here, $w^*_{\text{syn}}$ refers to the optimal hyperparameter obtained after LOOCV with proximity-based weighting as described in \Eqref{eq:weighting_function}, and the other weights follow the fixed scheme.}} 
\label{fig:sample_weight_brake}
\end{figure}

\begin{figure}[!htb]
\centering
\subfloat[Direct augmentation]{\label{fig:sample_weight_violin1_brake}\includegraphics[width=.5\linewidth]{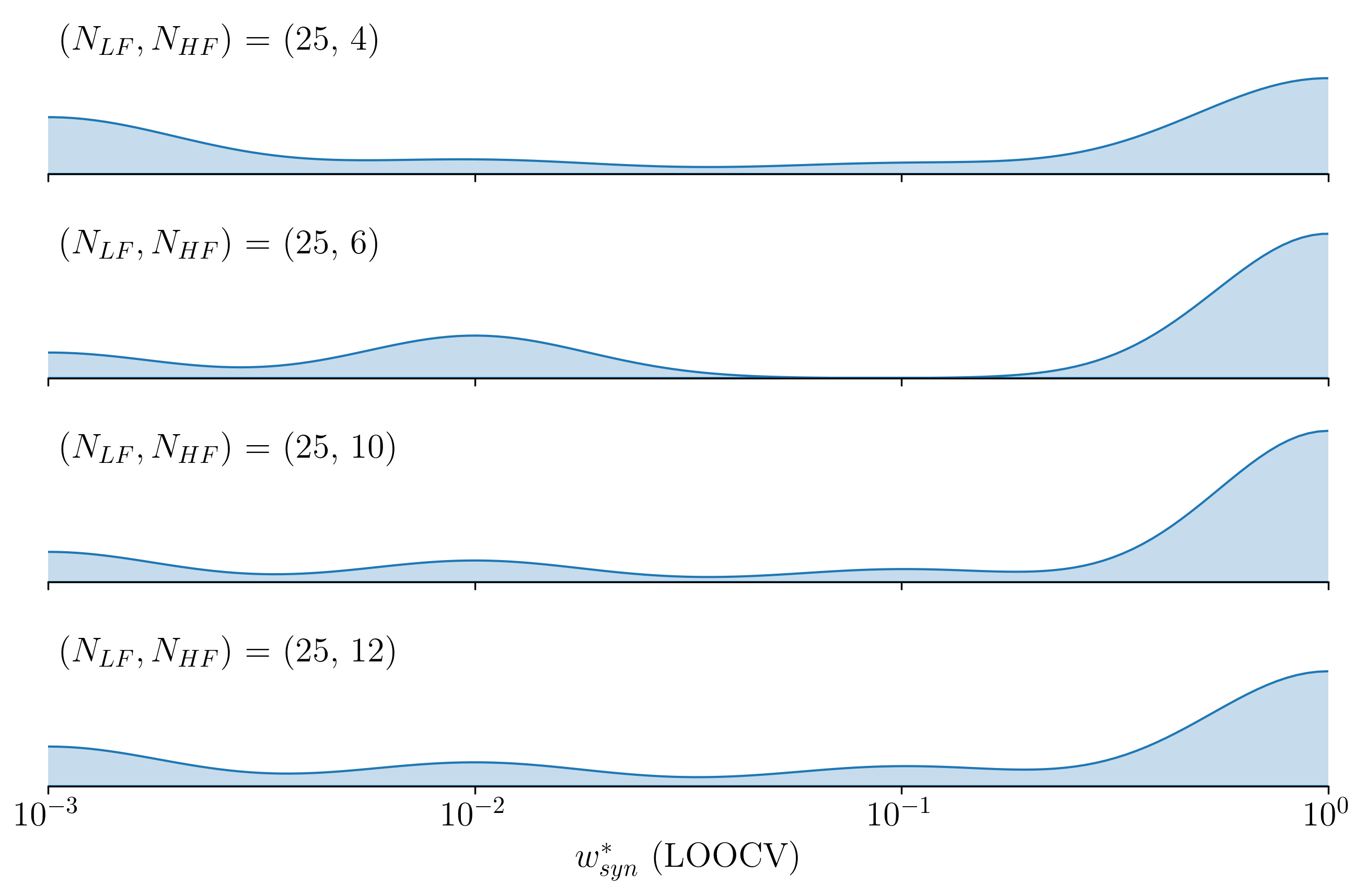}}\hfill
\subfloat[Explicit map]{\label{fig:sample_weight_violin2_brake}\includegraphics[width=.5\linewidth]{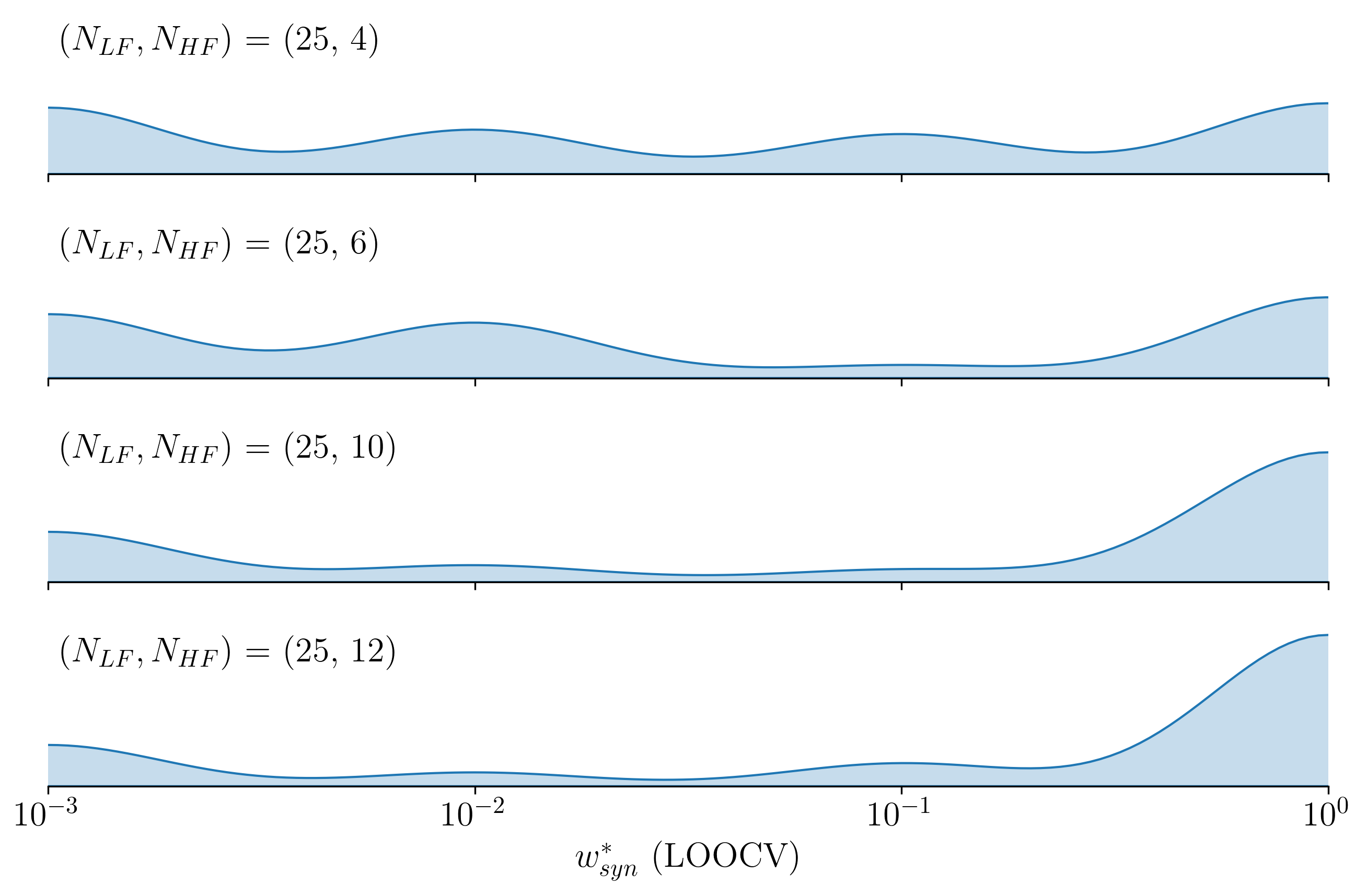}}
\caption{\new{Comparison of $w^*_{\text{syn}}$ distributions obtained from LOOCV for MF linear regression using data augmentation with proximity-based weighting on 50 repetitions of the training dataset for the aircraft brakes thermal modeling problem. Plotted are the kernel density estimates for each combination of LF and HF data.}}
\label{fig:sample_weight_violin_brake}
\end{figure}

\new{We begin the analysis by confirming that our tolerance $\epsilon = 0.9999$ yields a feasible number of dimensions in the reduced-space by observing the singular value decay and cumulative energy for both the LF and HF data. We can see from Figures~\ref{fig:LF_SVD_decay_brake}--\ref{fig:HF_SVD_decay_brake} that a reduced-dimension of size $k=4$ would be selected. In the following results and figures, the data augmentation based methods were fitted via a quadratic polynomial while the SF and additive based method had a linear fit. The best performing comparator was plotted for the fairest comparison, because utilizing a higher-order fit for the SF and additive based method resulted in a worse overall accuracy. This highlights the advantage of the WLS approach which allows for higher-order fits before being under-determined. The data in this problem led to both methods being largely invariant with respect to the choice of weighting as seen by Figure~\ref{fig:sample_weight_brake}. Importantly, the LOOCV procedure, which relies entirely on training data, still resulted in close to the best performing surrogate when robustness is taken into account. The distribution of the weighting hyperparameter $w_\text{syn}$ that were selected using LOOCV is shown in Figure~\ref{fig:sample_weight_violin_brake}. For each repetition of the training dataset, after the cross-validation procedure, weights close to $w^*_{\text{syn}}=1$ were selected most often. We note that the six-dimensional input space in this problem is greater than the three-dimensional input space in the hypersonics example, which drives the weights for the LF sample towards one for most case due to sparsity of available data. Thus, when fitting a quadratic (degree $p=2$) polynomial, any additional data would be prioritized by the LOOCV procedure during the weight selection. In this application, applying the explicit mapping initially led to a worse fit compared to direct DA with low HF samples, which again highlights the issues with learning an explicit map for a higher-dimensional input space with limited data. As more HF data was added, the learned map improves and the mapped data becomes more informative as reflected by better accuracy of the MF method. This can also be seen in the weight distribution plot in Figure~\ref{fig:sample_weight_violin2_brake} as smaller weights were picked more often than in the DA case.}

\begin{figure}[!htb]

\centering
\subfloat[Comparison of all methods]{\label{fig:mfcomparea_brake}\includegraphics[width=.5\linewidth]{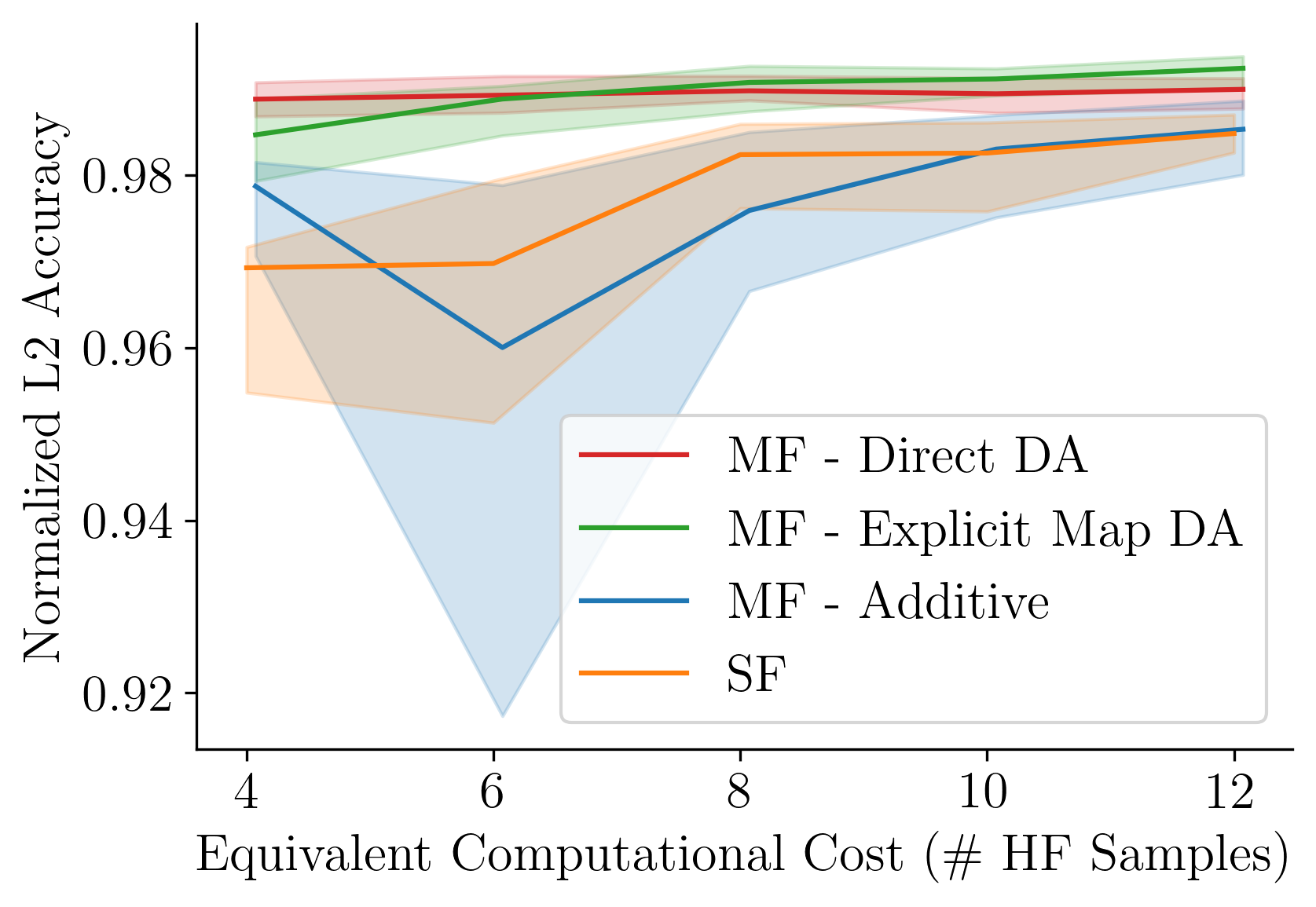}}\hfill
\subfloat[Additive method]{\label{mfcompareb_brake}\includegraphics[width=.5\linewidth]{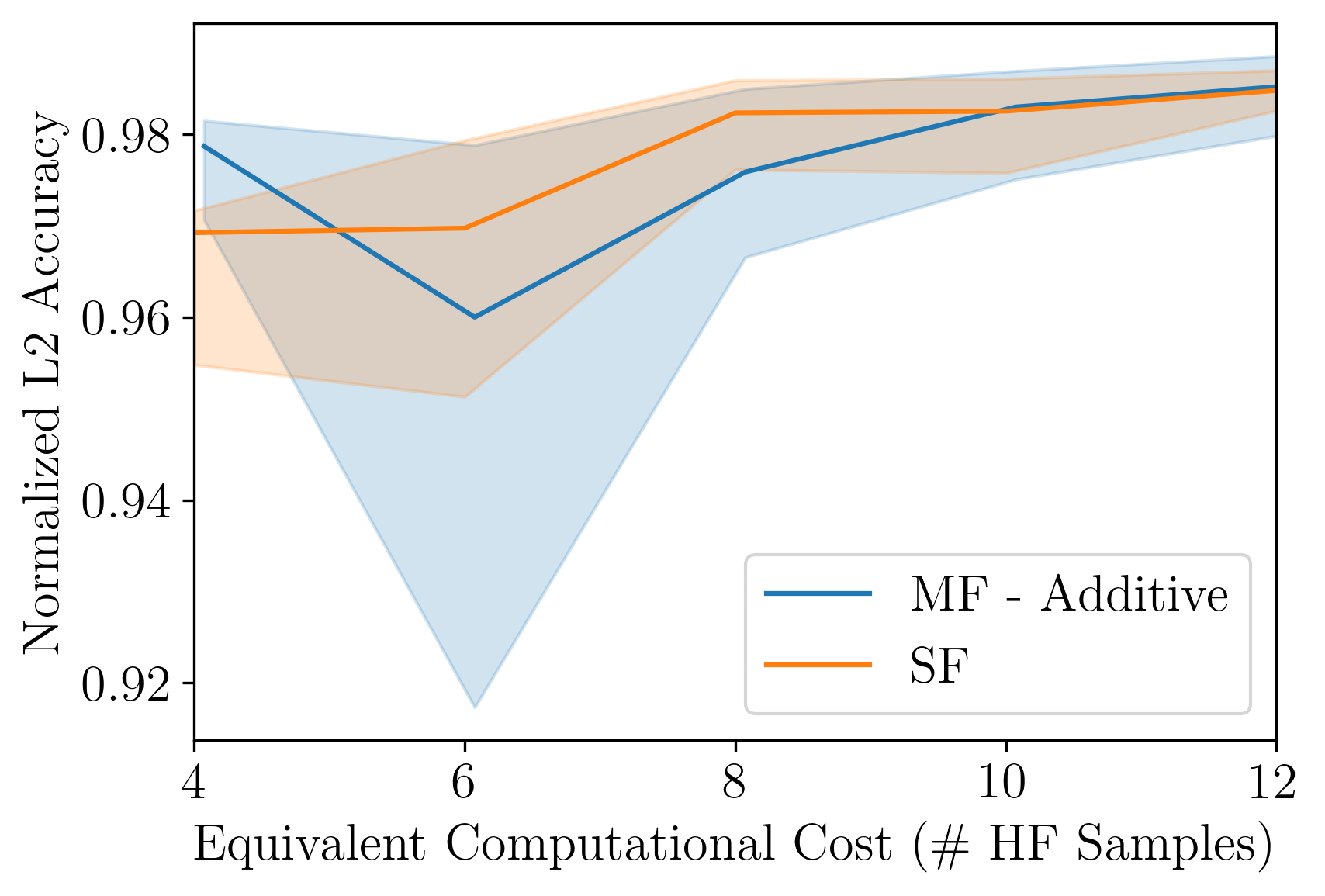}}\par 
\subfloat[Data augmentation: direct augmentation]{\label{mfcomparec_brake}\includegraphics[width=.5\linewidth]{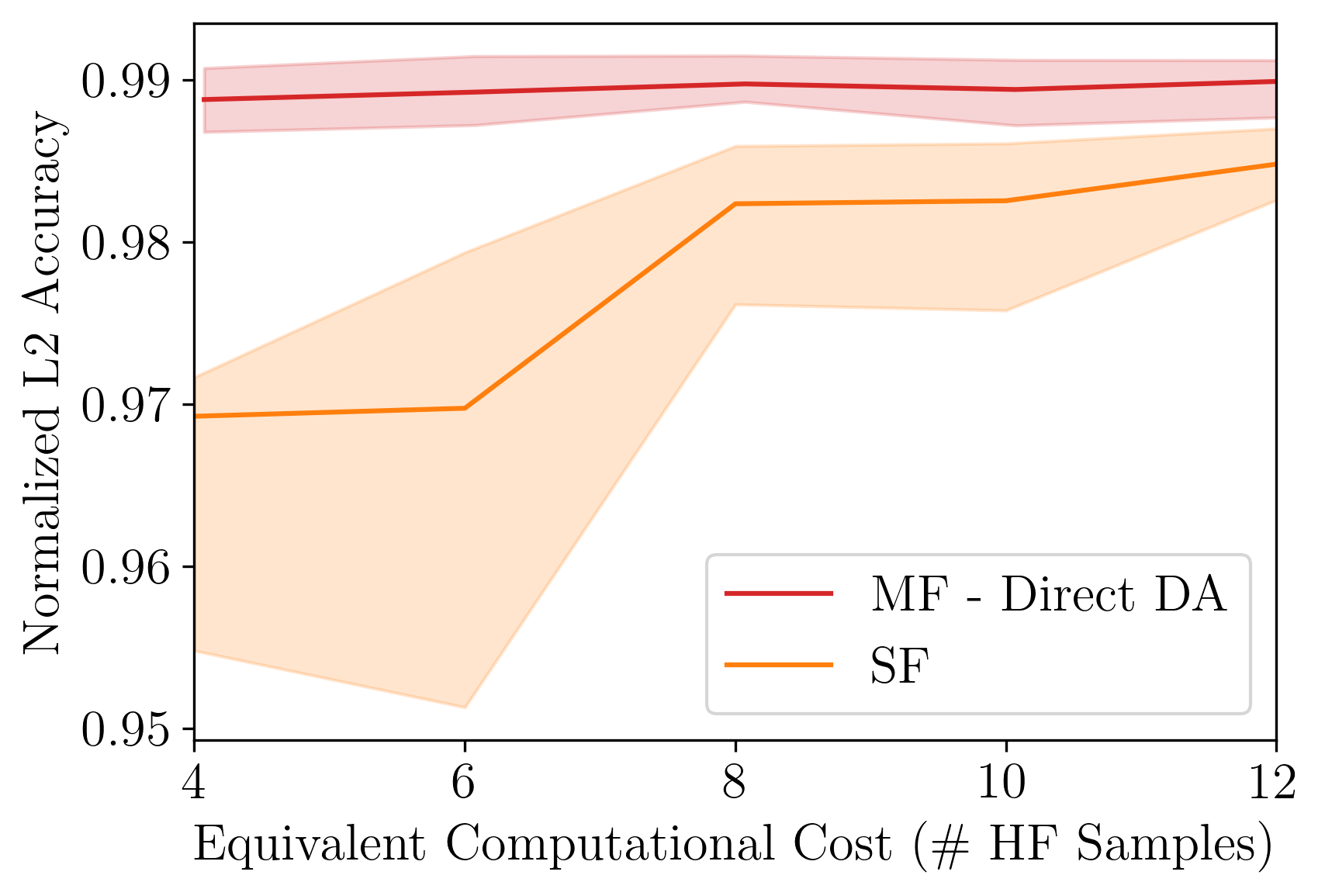}}\hfill
\subfloat[Data augmentation: explicit mapping]{\label{mfcompared_brake}\includegraphics[width=.5\linewidth]{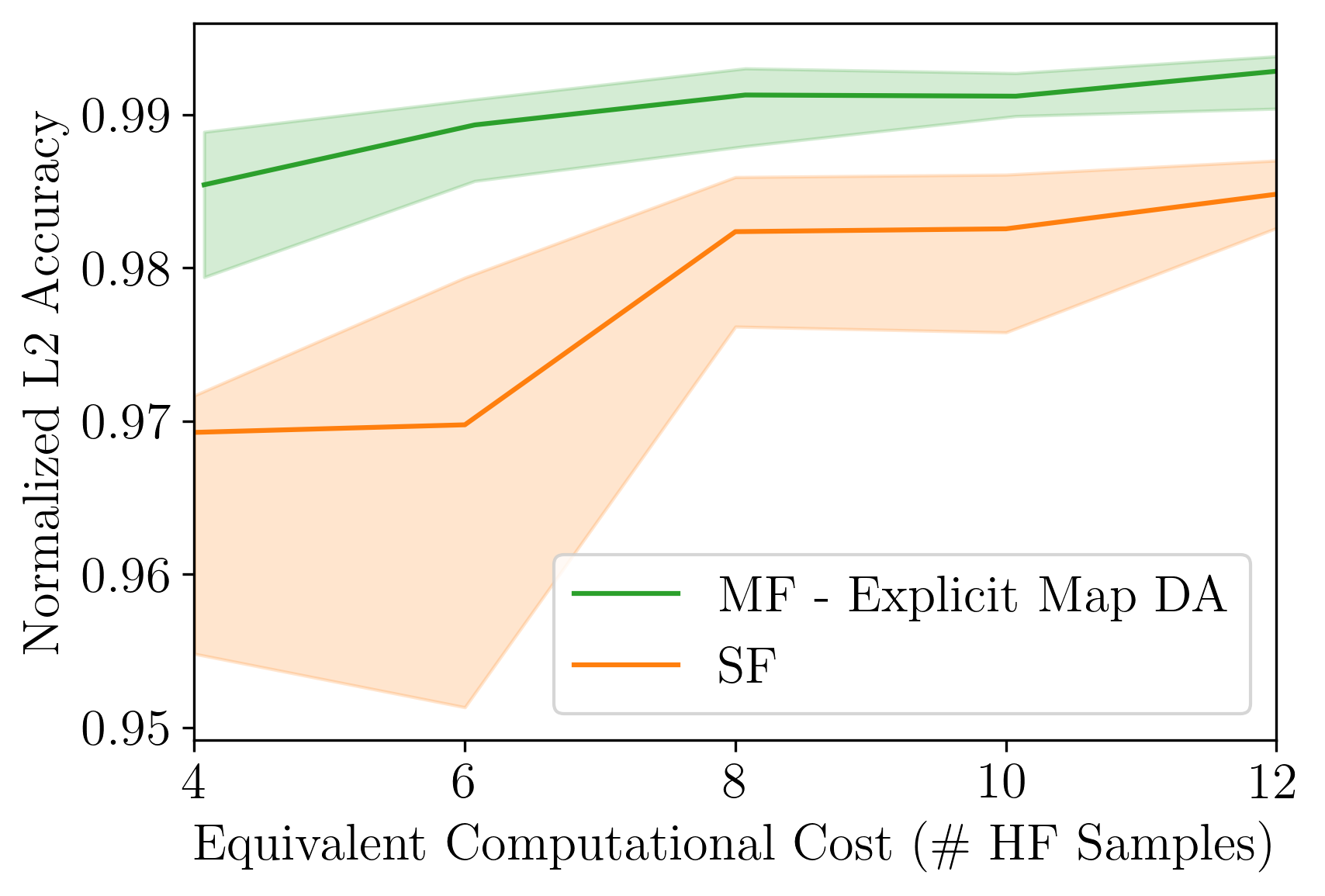}}
\caption{\new{Comparison of proposed MF linear regression methods to baseline SF linear regression and the additive MF method on 50 repetitions of the training dataset for the aircraft brake thermal modeling problem }}
\label{fig:mfcompare_brake}
\end{figure}

\new{Figure~\ref{fig:mfcompare_brake} shows the comparison of the two variants of the MF linear regression method proposed in this work with the additive MF method and the standard SF regression. The additive MF method performs similar to the SF linear regression and does not offer significant increase in accuracy for this application. In contrast, both the data augmentation techniques (using the optimal $w_\text{syn}$ after LOOCV and proximity-based weights) perform better than the additive approach and the SF linear regression for equivalent computational cost. Both the proposed MF techniques demonstrate more robustness to variations in training data with $\sim2\%$ improvement in the normalized L2 test accuracy. However, the MF linear regression methods show a significant improvement with respect to the $R^2$ score as seen from Figure~\ref{fig:mfcompare_brake_r2}. The remainder of the results utilizing $R^2$ values are again found in Appendix~\ref{sec:r2_results}. A selected summary of results for this application can be found in Table~\ref{tab:KOH_CART_table_brake}.}
\begin{table}[!htb]

\caption{\new{Selected multifidelity linear regression results  for the aircraft brake thermal modeling problem}}
    \centering
    \begin{tabularx}{\linewidth}{l*{4}{>{\centering\arraybackslash}X}}
        \toprule
        \textbf{Model Type} & \textbf{\# LF Samples} & \textbf{\# HF Samples} & \textbf{ Median Normalized L2 Test Accuracy} & \textbf{\new{Median Test Data $R^2$ Score}} \\ \midrule
        \multirow{3}{*}{SF} & - & 4 & 0.968 & 0.75 \\ 
         & - & 8 & 0.982 & 0.91 \\ 
         & - & 12 & 0.984 & 0.93 \\ \hline
        \multirow{3}{*}{MF - Additive} & \multirow{3}{*}{25} & 4 & 0.974 & 0.88 \\ 
         &  & 8 & 0.971 & 0.83\\
         &  & 12 & 0.978 & 0.94\\ \hline
         \multirow{3}{*}{\shortstack{MF - Direct DA (LOOCV $w^*_\text{syn}$)}} & \multirow{3}{*}{25} & 4 & 0.992 & 0.96\\ 
         &  & 8 & 0.994 & 0.98\\
         &  & 12 & 0.994 & 0.98\\ \hline
         \multirow{3}{*}{\shortstack{MF - Explicit map DA (LOOCV $w^*_\text{syn})$}} & \multirow{3}{*}{25} & 4 & 0.983 & 0.93\\ 
         &  & 8 & 0.993 & 0.97\\
         &  & 12 & 0.994 & 0.98\\ 
         \bottomrule
    \end{tabularx}
    \label{tab:KOH_CART_table_brake}
\end{table}

\begin{figure}[!htb]
	\centering
	\includegraphics[width=\textwidth]{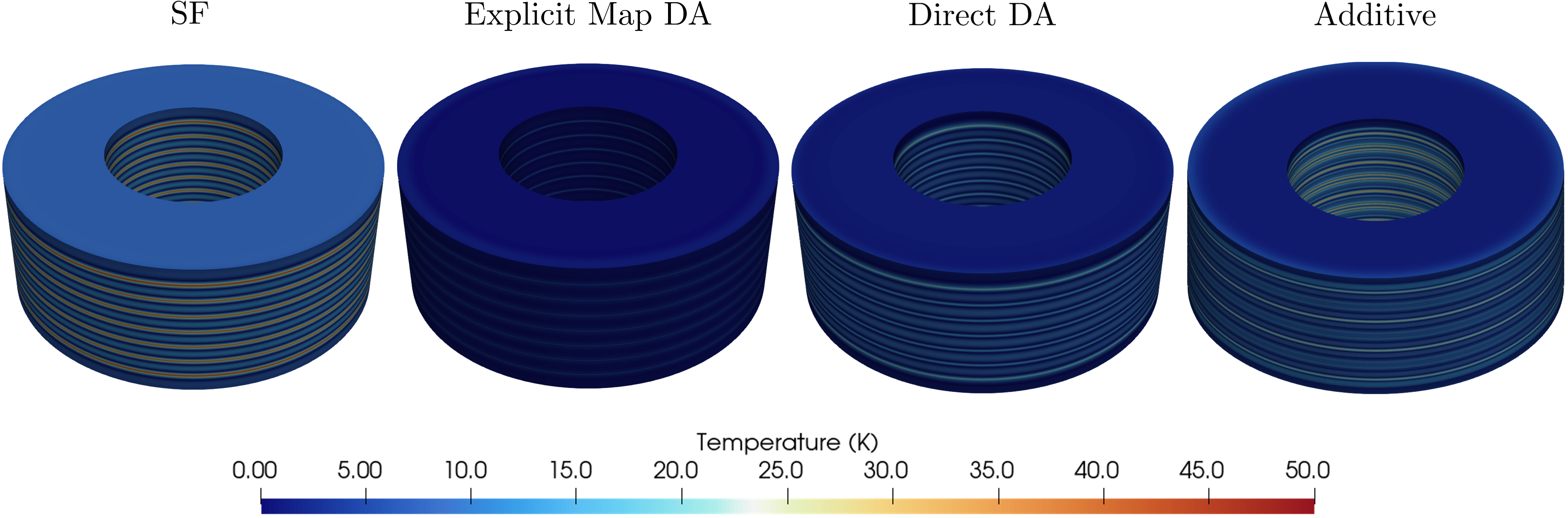}
	\caption{\new{Comparison of absolute errors in temperature field prediction at \textbf{Mass $=$ 84566 kg,} \textbf{Velocity $\boldsymbol{= 70.6}$ m/s,}
    $\boldsymbol{\tau_b = 27.1}$ \textbf{s,} $\boldsymbol{\alpha = 7.27\times10^{-6}}$ \textbf{m$^2$/s, } $\boldsymbol{T_\infty = 285}$ \textbf{K,} $\boldsymbol{T_\textbf{initial} = 441.6}$ \textbf{K}}
    }
	\label{fig:brakes_fig}
\end{figure}

\new{Lastly, we plot the absolute errors in temperature prediction between the SF surrogate and the MF surrogate methods. For an arbitrary test sample, we predict the temperature field using the surrogates and show the absolute error compared to the HF model simulation. We show a contour plot of the errors on the simplified cylindrical brake assembly in Figure \ref{fig:brakes_fig}, providing some visual context for the gains the MF surrogate model nets.}

\clearpage
\section{Conclusions}
\label{sec:conclusion}
This work presents MF linear regression methods \new{with linear and nonlinear features} for problems in the ultra low-data regime with two approaches using data augmentation. We embed dimensionality reduction through the principal component analysis with the MF regression methods to tackle high-dimensional outputs. As a point of comparison, we compare with the additive method for MF linear regression that uses the Kennedy O'Hagan framework with discrepancy function to correct the LF regression model. In the MF linear regression using data augmentation, we transform the LF data in two different ways and augment the transformed data to the HF dataset to perform a weighted least squares linear regression. The MF method uses proximity-based weighting strategy with cross-validation to \new{automatically} select the optimal weighting parameters. \new{To compare and contrast the MF approaches, two numerical examples are used: (1) the prediction of the pressure load on a hypersonic vehicle in-flight with a three-dimensional input space and (2) the prediction of the temperature field in aircraft disc-brakes after braking with a six-dimensional input space.} \new{For these applications with HF training samples} in the range of three to \new{twelve}, we find that the data augmentation techniques with proximity-based weighting produce robust and accurate surrogate models leading to approximately $2-12\%$ in median accuracy gain in the low-data regime as compared to the SF surrogate. The additive approach does not substantially improve the accuracy compared to the baseline SF surrogate model. The direct data augmentation method had comparable accuracy to the explicit mapping method \new{for both examples}, but showed more sensitivity to the selection of the synthetic data weight \new{when applied to the hypersonics testbed problem}. Both direct data augmentation and explicit mapping methods work robustly and accurately across variations in training data when used with proximity-based weighting and automatic weight selection through cross-validation.
\new{The proposed MF linear regression methods demonstrated the capability to robustly predict the high-dimensional quantities of interest in both numerical examples. 
}

Future work can expand these MF regression methods to different underlying regression techniques, such as neural networks and regression trees. Another research direction would be to explore different coordinate transformation techniques for the explicit mapping method.

\section*{Declarations}
\subsection*{Funding}
This work has been supported in part by ARPA-E Differentiate award number DE-AR0001208, AFOSR grant FA9550-21-1-0089 under the NASA University Leadership Initiative (ULI), AFOSR grant FA9550-24-1-0327 under the Multidisciplinary University Research Initiatives (MURI), DARPA Automating Scientific Knowledge Extraction and Modeling (ASKEM) program award number DE-AC05-76RL01830, DARPA The Right Space (TRS) award number HR00112590114, and DOE ASCR grant DE-SC002317.

\begin{appendices}

\section{Multifidelity linear regression with an additive structure}\label{app:mf-add}
We \new{use the state-of-the-art} additive MF linear regression method based on the Kennedy–O’Hagan framework~\cite{KOH_seminal}, adapted to operate on reduced-order representations of the outputs \new{to provide a fair comparison with our proposed approach}. The method in Ref.~\cite{zhang2018multifidelity} is equipped with dimensionality reduction of the outputs and is used as a point of comparison for the data-augmentation-based MF methods proposed in this work.
This method chooses to model the relationship between the LF and the HF data linearly and needs co-located data to estimate discrepancy by HF and LF models. 
The first component is a LF surrogate model $f_\text{LF}$, trained on the reduced LF outputs given by \Eqref{eq:projection_down}, using OLS on the dataset $\left(\boldsymbol{X}_\text{LF}, \boldsymbol{C}_\text{LF}\right)$. 
Similar to the explicit mapping method (see Section~\ref{sec:DA}), to obtain co-located LF output predictions, the predictions of reduced LF states at the HF input locations $\boldsymbol{X}_\text{HF}$ are obtained by $f_\text{LF}(\boldsymbol{X}_{\text{HF}})$ and reconstructed to the full-dimensional output space as $\boldsymbol{U}_k^\text{LF} f_\text{LF}(\boldsymbol{X}_{\text{HF}}) + \overline{\boldsymbol{Y}}_{\text{LF}}$.
The discrepancy data between the HF outputs and the co-located LF predictions is then computed as
\begin{equation}
    \label{eq:discrepancy}
  \boldsymbol{\delta}(\boldsymbol{X}_{\text{HF}}) = \boldsymbol{Y}_{\text{HF}} - (\boldsymbol{U}_k^\text{LF} f_\text{LF}(\boldsymbol{X}_{\text{HF}}) + \overline{\boldsymbol{Y}}_{\text{LF}}).
\end{equation}
A second surrogate model $f_\delta$ is trained via OLS on the reduced discrepancy data $\left(\boldsymbol{X}_\text{HF}, \boldsymbol{C}_{\boldsymbol{\delta}} \right)$ obtained through \Eqref{eq:discrepancy} and \Eqref{eq:projection_down}. Then, the predictions from the additive MF regression model in the full-dimensional space at any new input location $\boldsymbol{x}^*$ is given by
\begin{equation}
    \label{eq:mfadd}
    \begin{split}
    \widehat{y}_{\text{MF}}(\boldsymbol{x}^*) &= \boldsymbol{U}_k^\text{LF} f_\text{LF}(\boldsymbol{x}^*) + \overline{\boldsymbol{Y}}_{\text{LF}} + \boldsymbol{U}_k^{\delta}f_\delta(\boldsymbol{x}^*) + \overline{\boldsymbol{\delta}} \\
    &=  \underbrace{\boldsymbol{U}_k^\text{LF} f_\text{LF}(\boldsymbol{x}^*)}_\text{LF model} + \underbrace{\boldsymbol{U}_k^{\delta}f_\delta(\boldsymbol{x}^*)}_\text{discrepancy model} + \underbrace{(\overline{\boldsymbol{Y}}_{\text{LF}} + \overline{\boldsymbol{\delta}})}_\text{bias},
    \end{split}
\end{equation}
where $\boldsymbol{U}_k^{\delta}$ is the reduced basis obtained via PCA on the discrepancy data and $\overline{\boldsymbol{\delta}}$ is the sample mean. The procedure for the projection-based additive MF regression model is summarized in Alg.~\ref{alg:add_arch}.

\begin{algorithm}[!htb]
\caption{Multifidelity linear regression via an additive method} \label{alg:add_arch}
\begin{algorithmic}[1]
\Statex \textbf{Input:} HF and LF training data $(\boldsymbol{X}_{\text{LF}}, \, \boldsymbol{Y}_{\text{LF}})$ and $(\boldsymbol{X}_{\text{HF}}, \, \boldsymbol{Y}_{\text{HF}})$, new input locations for prediction $\boldsymbol{x}^*$ 
\Statex \textbf{Output:} Output predictions $\widehat{\boldsymbol{y}}_{\text{MF}}(\boldsymbol{x}^*)$ at input location $\boldsymbol{x}^*$ from MF surrogate
\State Project $\boldsymbol{Y}_{\text{LF}}$ to obtain the reduced states $\boldsymbol{C}_\text{LF}=\left(\boldsymbol{U}_k^\text{LF}\right)^\top \left(\boldsymbol{Y}_{\text{LF}} - \overline{\boldsymbol{Y}}_{\text{LF}}\right)$ \Comment{see \Eqref{eq:projection_down}}
\State Train LF linear regression model $f_{\text{LF}}$ on $(\boldsymbol{X}_{\text{LF}},\boldsymbol{C}_\text{LF})$ using OLS
\State Predict and reconstruct LF outputs at the HF input locations $(\boldsymbol{U}_k^\text{LF} f_\text{LF}(\boldsymbol{X}_{\text{HF}}) + \overline{\boldsymbol{Y}}_{\text{LF}})$
\State Estimate discrepancy data $\boldsymbol{\delta}(\boldsymbol{X}_{\text{HF}})$ using \Eqref{eq:discrepancy}
\State Use $\boldsymbol{U}_k^{\delta}$ from the SVD of $\boldsymbol{\delta}$ to project the discrepancy to the reduced state $\boldsymbol{{C}}_{\delta} = (\boldsymbol{U}_k^{\delta})^\top (\boldsymbol{\delta} - \overline{\boldsymbol{\delta}})$ 
\State Train discrepancy linear regression model $f_{\delta}$ on ($\boldsymbol{X}_{\text{HF}}, \, \boldsymbol{{C}}_{\delta}$) using OLS
\State Predict outputs $\widehat{\boldsymbol{y}}_{\text{MF}}(\boldsymbol{x}^*)$ at new input location $\boldsymbol{x}^*$ as the linear combination of $f_{\delta}$, $f_{\text{LF}}$, and the known bias terms using \Eqref{eq:mfadd}
\end{algorithmic}
\end{algorithm}

\section{{Multifidelity results using the R$^2$ score}}
\label{sec:r2_results}
\begin{figure}[h]
\centering
\subfloat[Comparison of all methods]{\label{fig:mfcomparea_r2}\includegraphics[width=.5\linewidth]{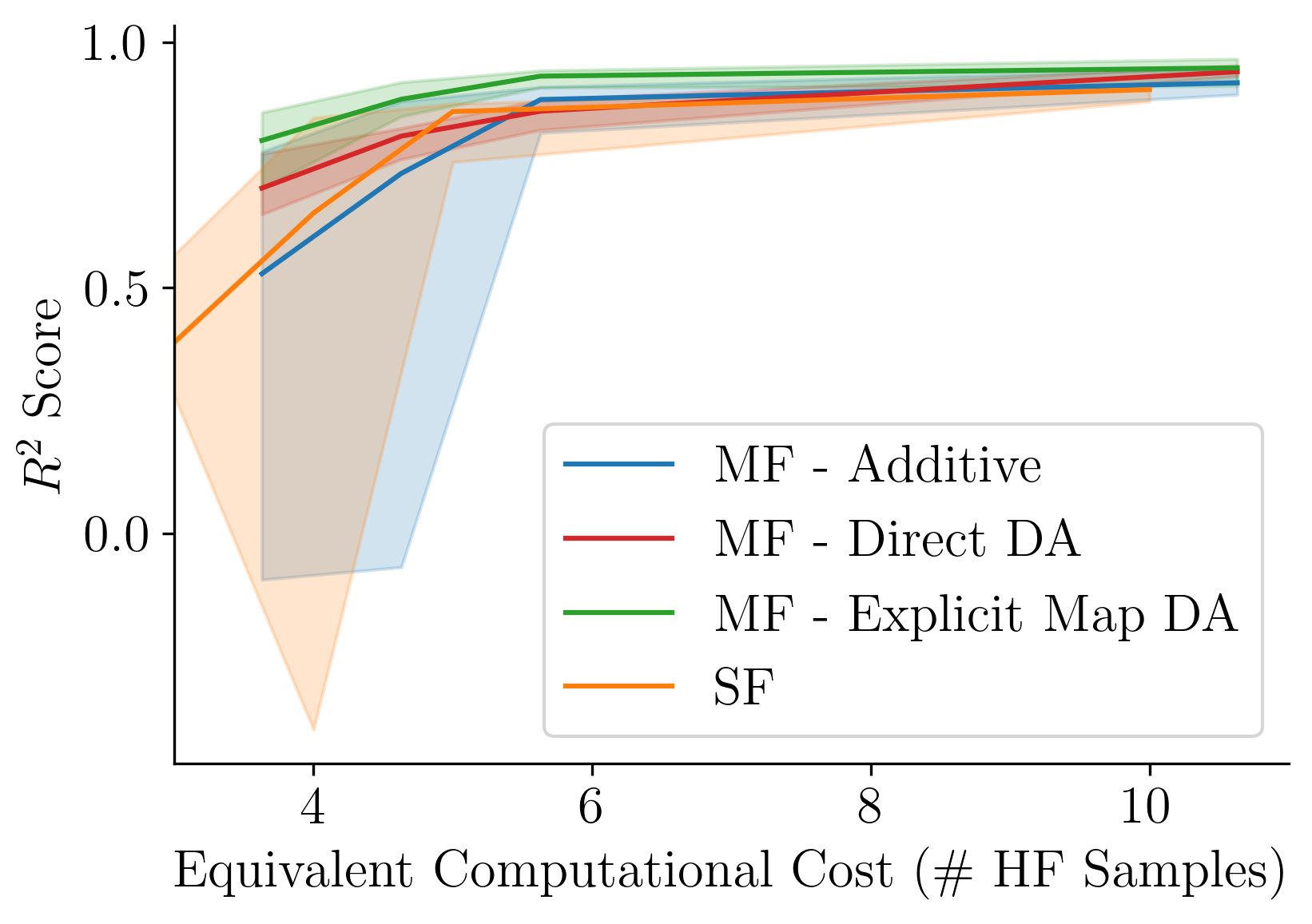}}\hfill
\subfloat[Additive method]{\label{mfcompareb_r2}\includegraphics[width=.5\linewidth]{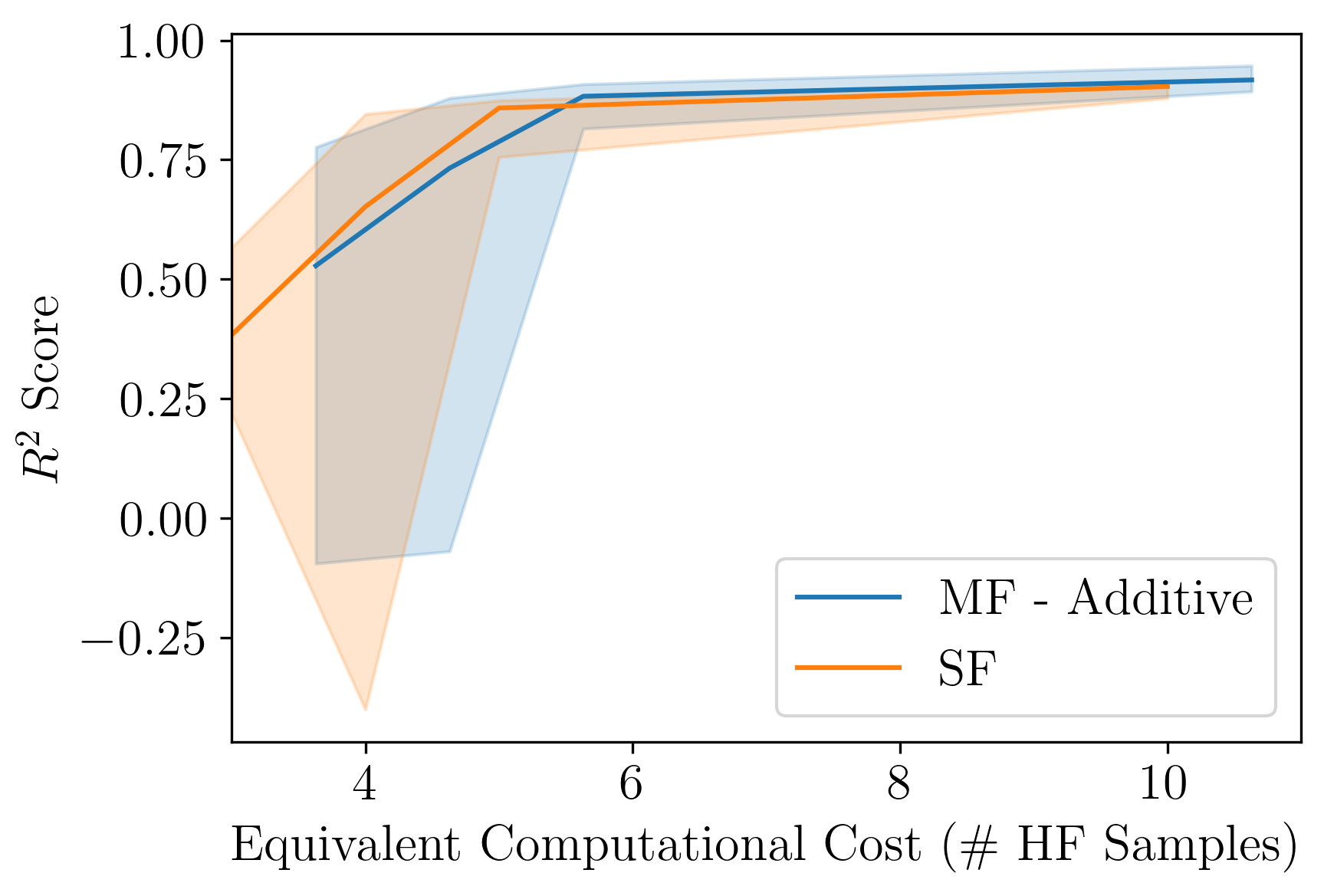}}\par 
\subfloat[Data augmentation: direct augmentation]{\label{mfcomparec_r2}\includegraphics[width=.5\linewidth]{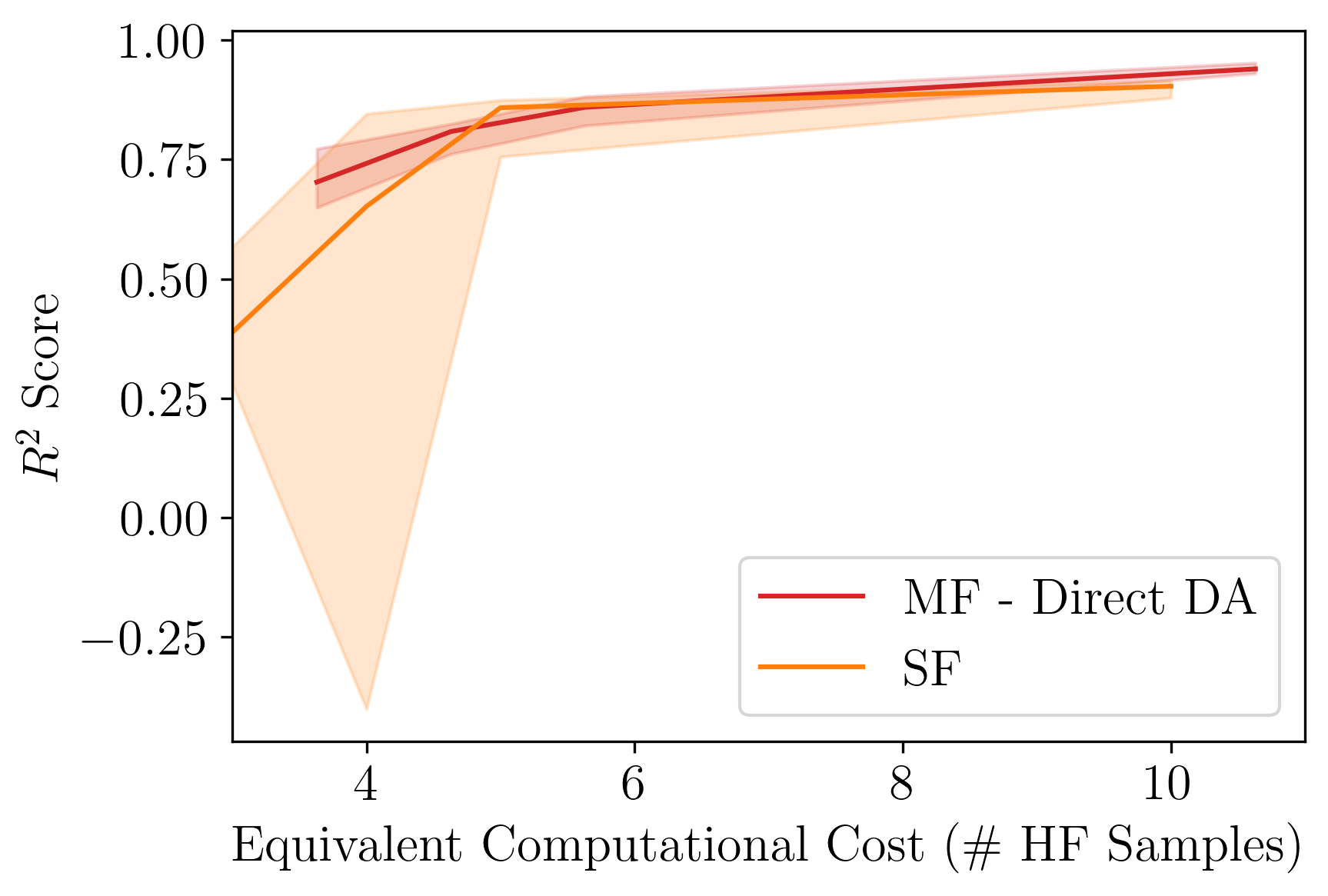}}\hfill
\subfloat[Data augmentation: explicit mapping]{\label{mfcompared_r2}\includegraphics[width=.5\linewidth]{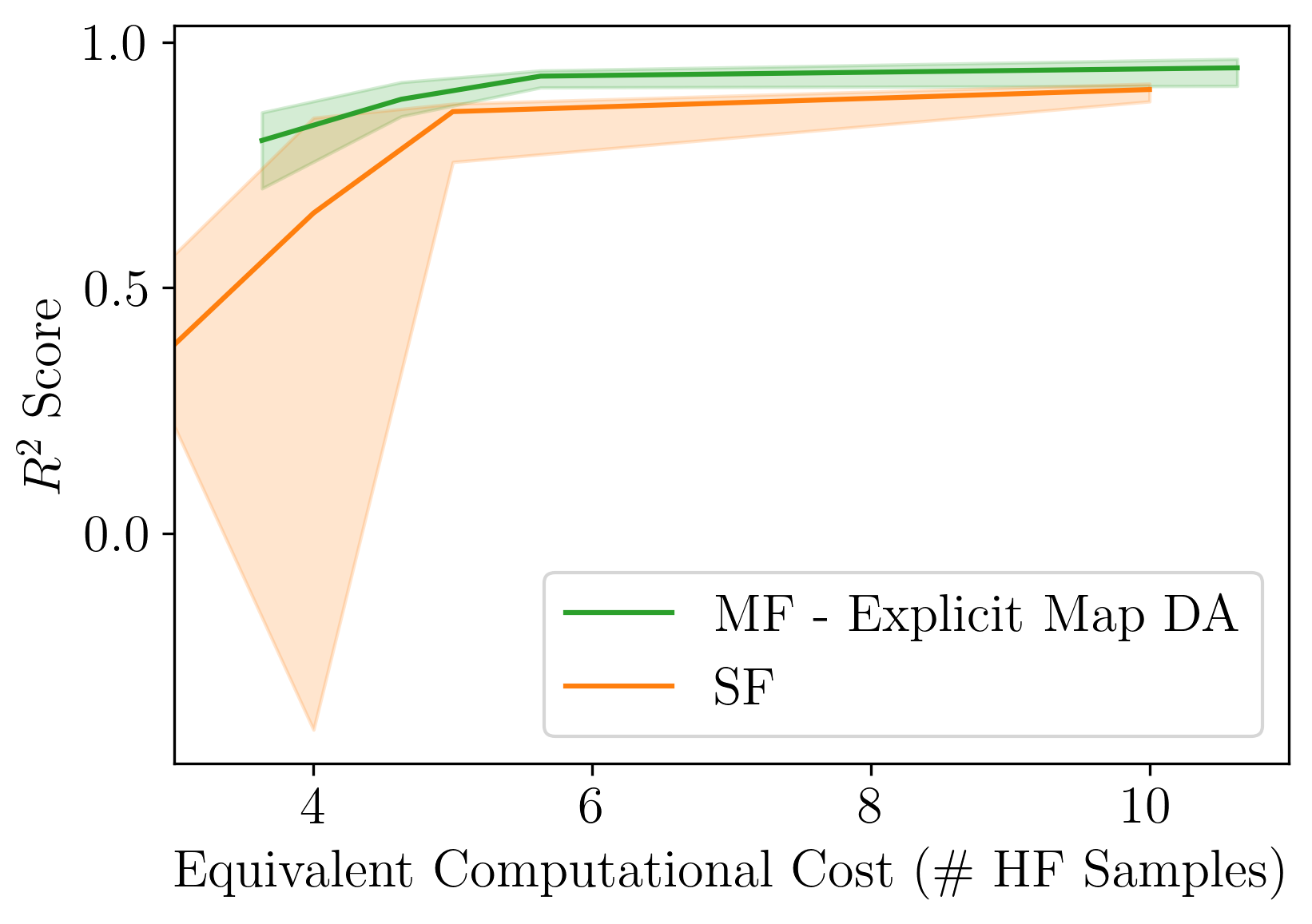}}
\caption{\new{Comparison of proposed MF linear regression methods to baseline SF linear regression and the additive MF method on 50 repetitions of the training dataset via the R$^2$ score for the hypersonic testbed problem}}
\label{fig:mfcompare_r2}
\end{figure}

\begin{figure}[h]
\centering
\subfloat[Comparison of all methods]{\label{fig:mfcomparea_brake_r2}\includegraphics[width=.5\linewidth]{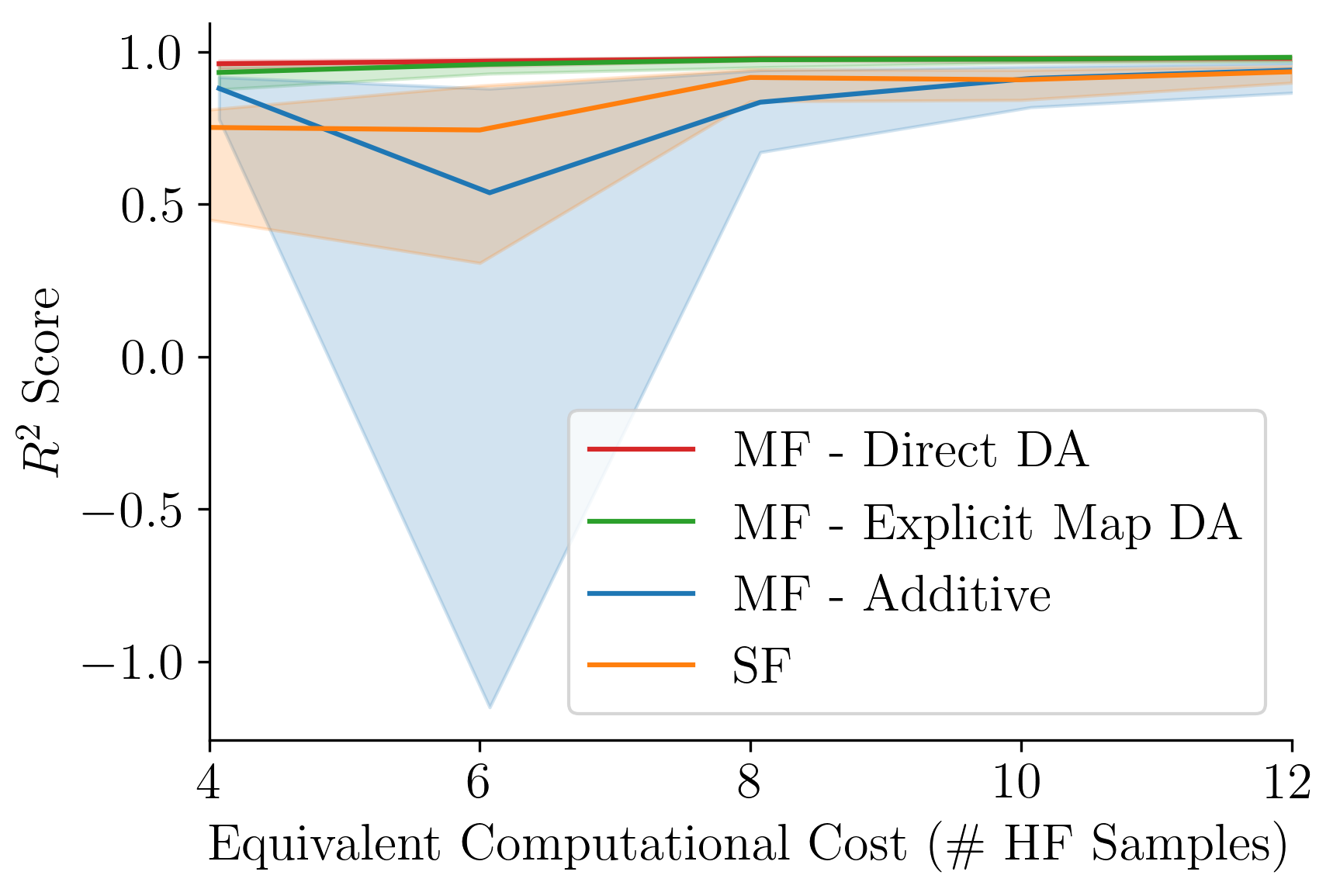}}\hfill
\subfloat[Additive method]{\label{mfcompareb_brake_r2}\includegraphics[width=.5\linewidth]{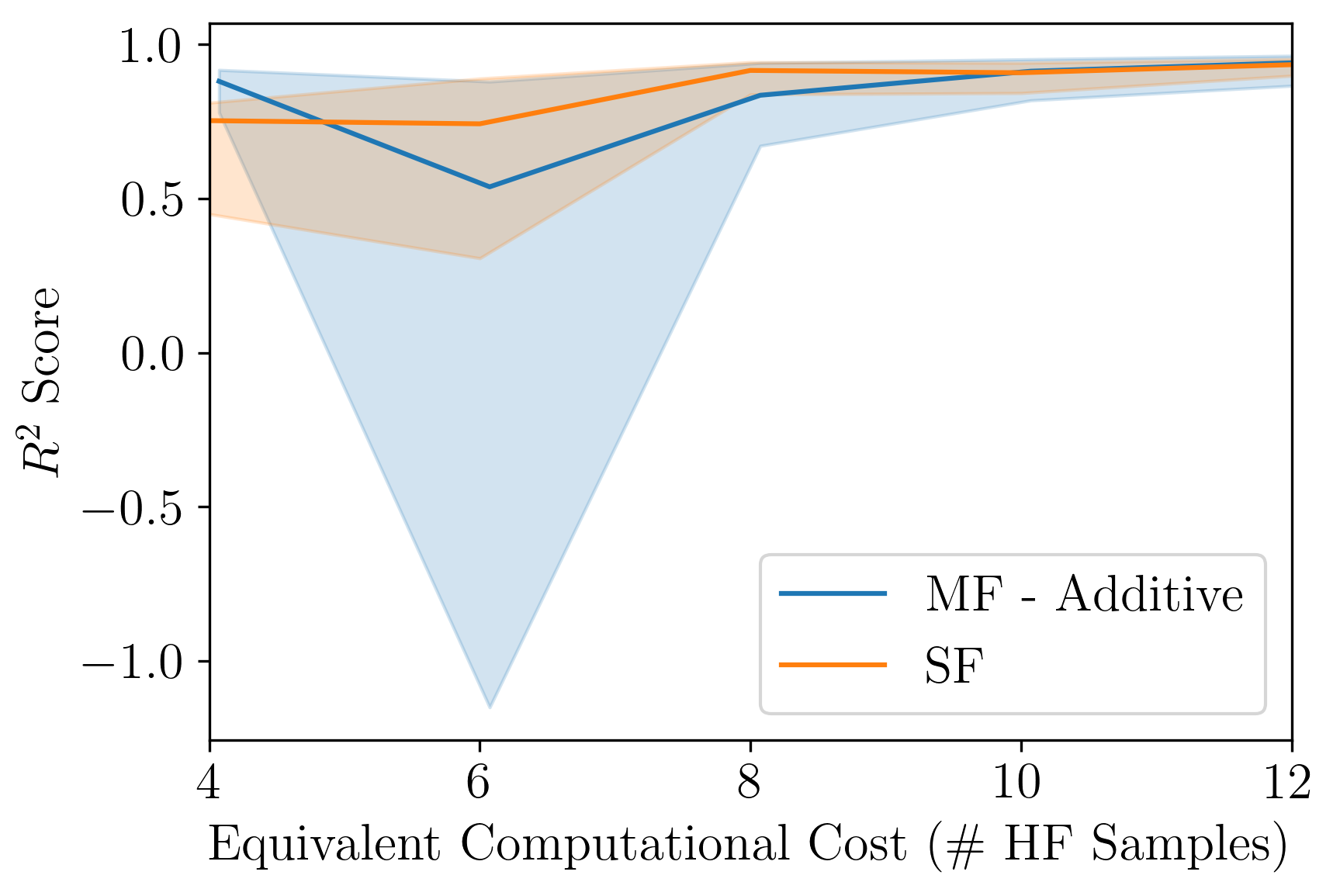}}\par 
\subfloat[Data augmentation: direct augmentation]{\label{mfcomparec_brake_r2}\includegraphics[width=.5\linewidth]{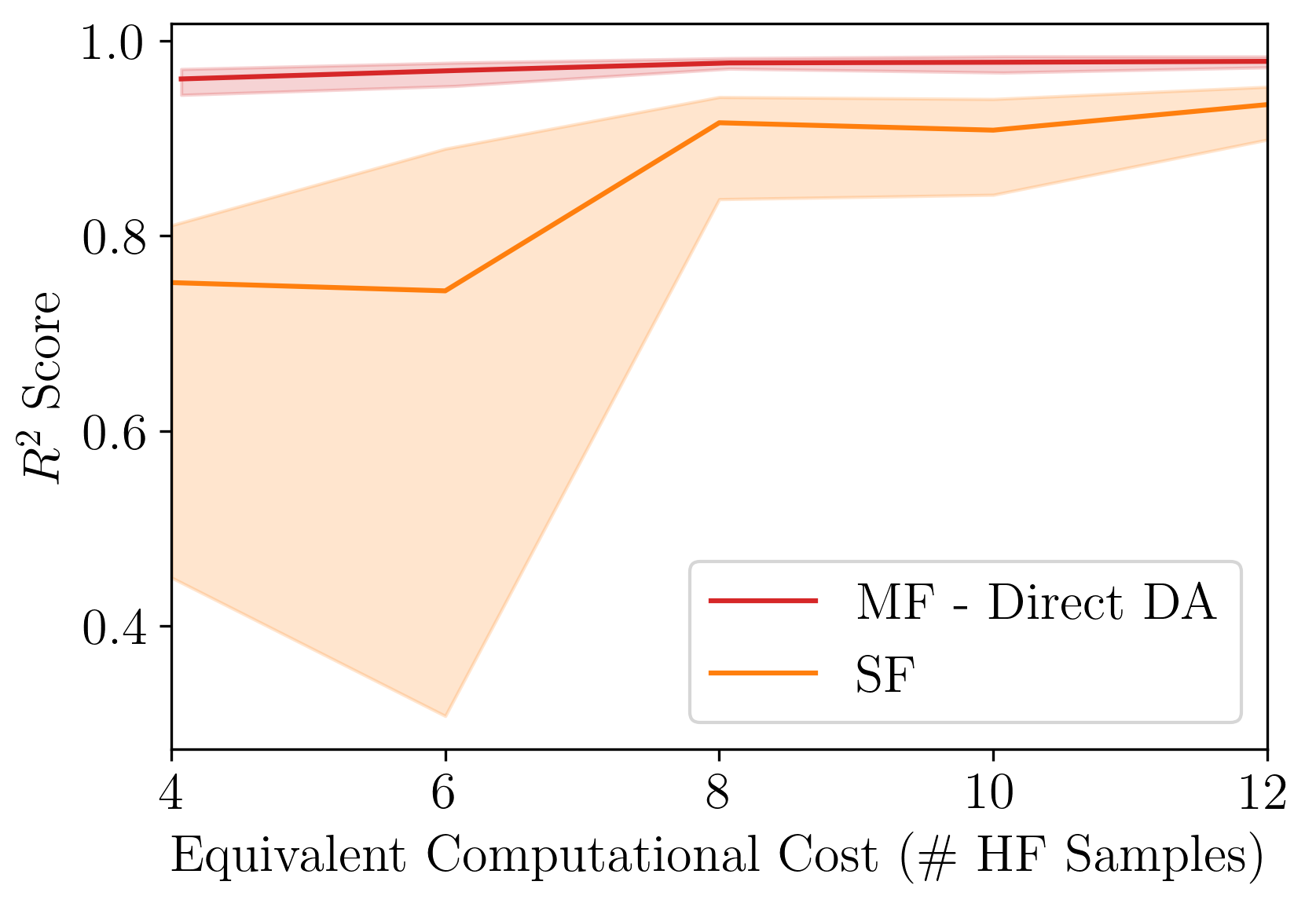}}\hfill
\subfloat[Data augmentation: explicit mapping]{\label{mfcompared_brake_r2}\includegraphics[width=.5\linewidth]{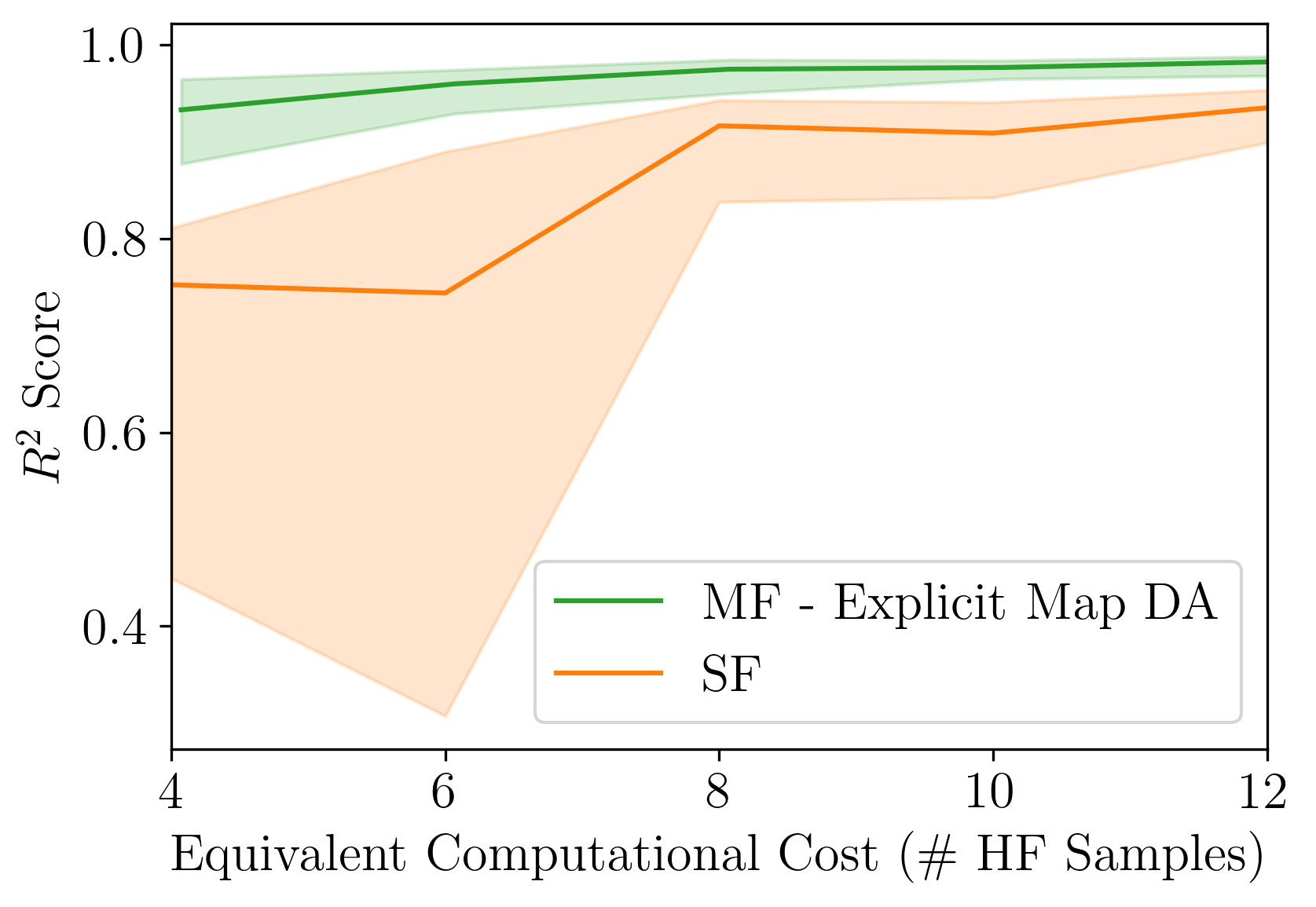}}
\caption{\new{Comparison of proposed MF linear regression methods to baseline SF linear regression and the additive MF method on 50 repetitions of the training dataset via the R$^2$ score for the aircraft brake thermal modeling problem}}
\label{fig:mfcompare_brake_r2}
\end{figure}

\new{We also present the results showing the coefficient of determination ($R^2$ score) in Figure~\ref{fig:mfcompare_r2} for the hypersonics problem and Figure~\ref{fig:mfcompare_brake_r2} for the aircraft disc brake problem. We see from Figure~\ref{mfcompared_r2} that the explicit mapping method had the best $R^2$ for the hypersonics example and was more robust to changes in the training data according to this particular criterion. The explicit mapping and direct augmentation methods performed similarly for the aircraft disc brake application.}

\section{{Conservation of energy for the heat source term}}\label{app:conserve_source}
\new{We show that the source term used to implement the frictional heating at the interfaces of the brake discs conserves the correct amount of energy. When an aircraft is braking, we assume the brakes will only have to dissipate the initial kinetic energy of the aircraft and no other sources. Thus, the total energy of the system at time $t=0$ is $E_\text{total} = \frac{1}{2}mV^2$ where $m$ is the mass of the aircraft (kg) and $V$ is the velocity (m/s).  Given that there are four braked wheels on the aircraft considered in this work, we define $W= 4$ to represent the number of brake stacks. Each brake stack contains $N$ discs resulting in the energy needed to be dissipated by one stack as}
\new{
\begin{align}
    E_\text{stack} = (N-1)\tau_bq_kA_\text{face},
    \label{eq:e_stack}
\end{align}
where $N-1$ is the number of interfaces between the $N$ discs, $\tau_b$ (s) is the braking period, $A_\text{face}$ (m$^2$) is the surface area of a rotor-stator interface, and $q_k$ (W/m$^2$) is the heat flux as defined later in \Eqref{eq:qk_term}.  The total number of interfaces between brake discs undergoing heating is given by $W(N-1)$. In our frictional heating source term in \Eqref{eq:heat_source}, we define each $q_k$ such that the total energy of the aircraft is correctly applied within each stack with no variation over time as}
\new{
\begin{align}
   q_k = \frac{E_\text{total}}{W(N-1)A_\text{face}\tau_b}.
   \label{eq:qk_term}
\end{align}}

\new{This formulation will result in the correct amount of power, and therefore energy, being applied at each interface. The power required to arrest the brakes is given by the kinetic energy being dissipated as heat through the heat source $f(r,z,t)$ as
\begin{align*} 
P_{\text{interface}} &= \int_{r_i}^{r_o} \int_{z_k-\varepsilon/2}^{z_k+\varepsilon/2} f(r,z,t)  2\pi r  \,\textrm{d}z  \,\textrm{d}r \\
&= \int_{r_i}^{r_o} \int_{z_k-\varepsilon/2}^{z_k+\varepsilon/2} \dfrac{q_k}{\varepsilon}  2\pi r  \,\textrm{d} z  \,\textrm{d}r \\
&= \int_{r_i}^{r_o} q_k  2\pi r  \,\textrm{d}r \\
& = q_k  A_{\text{face}}.
\end{align*}
Thus, the energy dissipated over $N-1$ interfaces in a stack over time $\tau_b$ is
\begin{align*} 
(N-1)\int_{0}^{\tau_b} P_{\text{interface}}  dt
= (N-1)\tau_b  q_k  A_{\text{face}}.
\end{align*}
which matches the expression for $E_\text{stack}$ in \Eqref{eq:e_stack}.}

\end{appendices}

\bibliography{sn-bibliography}%

\end{document}